\documentclass[10pt,journal, compsoc]{IEEEtran}
\ifCLASSOPTIONcompsoc
\usepackage[nocompress]{cite}
\usepackage{amsmath,graphicx,mathrsfs,booktabs,type1cm,amssymb,threeparttable,multirow, booktabs, changepage, subfigure}
\else
\usepackage{cite}
\usepackage{amsmath,graphicx,mathrsfs,booktabs,type1cm,amssymb,threeparttable,multirow, booktabs, changepage, subfigure}
\fi
\usepackage{footnote}
\usepackage{color}
\usepackage{verbatim}
\usepackage{bm}
\usepackage{makecell, rotating}
\usepackage[colorlinks,linkcolor=blue]{hyperref}

\ifCLASSINFOpdf
\else
\fi

\begin{document}
	
	\title{GCNet: Graph Completion Network for Incomplete Multimodal Learning in Conversation}

	\author{Zheng~Lian,~\IEEEmembership{}
		Lan~Chen,~\IEEEmembership{}
		Licai~Sun,~\IEEEmembership{}
		Bin~Liu,~\IEEEmembership{Member,~IEEE,}
		and~Jianhua~Tao,~\IEEEmembership{Senior Member,~IEEE} % <-this % stops a space

		% <-this % stops a space
		\IEEEcompsocitemizethanks{
			
			\IEEEcompsocthanksitem Zheng Lian, Lan Chen and Bin Liu are with National Laboratory of Pattern Recognition, Institute of Automation, Chinese Academy of Sciences, Bejing, China, 100190.
			E-mail: lianzheng2016@ia.ac.cn; chenlan2016@ia.ac.cn; liubin@nlpr.ia.ac.cn.
			\protect
			
			\IEEEcompsocthanksitem Licai Sun is with the School of Artificial Intelligence, University of Chinese Academy of Sciences, Beijing, China, 100049.
			E-mail: sunlicai2019@ia.ac.cn.
			\protect
			
			\IEEEcompsocthanksitem Jianhua Tao is with Department of Automation, Tsinghua University, Bejing, China, 100084 and with the School of Artificial Intelligence, University of Chinese Academy of Sciences, Beijing, China, 100049.
			E-mail: jhtao@tsinghua.edu.cn.
			\protect
		}
		
		% <-this % stops an unwanted space
		\thanks{Manuscript received \(\rm{xxxxxxxx}\); revised \(\rm{xxxxxxxx}\). (Corresponding author: Jianhua Tao, Bin Liu)}
	}

	\markboth{IEEE TRANSACTIONS ON PATTERN ANALYSIS AND MACHINE INTELLIGENCE}%
	{Shell \MakeLowercase{\textit{et al.}}: Bare Demo of IEEEtran.cls for Computer Society Journals}

	
	\IEEEtitleabstractindextext{%
		\begin{abstract}
			Conversations have become a critical data format on social media platforms. Understanding conversation from emotion, content and other aspects also attracts increasing attention from researchers due to its widespread application in human-computer interaction. In real-world environments, we often encounter the problem of incomplete modalities, which has become a core issue of conversation understanding. To address this problem, researchers propose various methods. However, existing approaches are mainly designed for individual utterances rather than conversational data, which cannot fully exploit temporal and speaker information in conversations. To this end, we propose a novel framework for incomplete multimodal learning in conversations, called ``Graph Complete Network (GCNet)'', filling the gap of existing works. Our GCNet contains two well-designed graph neural network-based modules, ``Speaker GNN'' and ``Temporal GNN'', to capture temporal and speaker dependencies. To make full use of complete and incomplete data, we jointly optimize classification and reconstruction tasks in an end-to-end manner. To verify the effectiveness of our method, we conduct experiments on three benchmark conversational datasets. Experimental results demonstrate that our GCNet is superior to existing state-of-the-art approaches in incomplete multimodal learning. \textcolor[rgb]{0.93,0.0,0.47}{Code is available at https://github.com/zeroQiaoba/GCNet.}
		\end{abstract}
		
		}
	
	\maketitle
	\IEEEdisplaynontitleabstractindextext
	\IEEEpeerreviewmaketitle
	
	\IEEEraisesectionheading{\section{Introduction}\label{sec:introduction}}
	\IEEEPARstart{C}{onversation} understanding has become an active research area due to its widespread applications in many tasks, including dialogue systems \cite{liang2022emotional, fu2022learning} and recommender systems \cite{nie2019multimodal, gao2021advances}. To understand conversations from different aspects (such as emotion and content), researchers collect a large amount of conversational data through various approaches, especially from social media platforms \cite{zadeh2016multimodal, zadeh2018multimodal}. However, in real-world environments, many factors may lead to missing modalities. For example, the speech is probably missing due to background noise or sensor failure; the text is perhaps unavailable due to automatic speech recognition errors or unknown words; the faces may not be detected due to lighting, motion, or occlusion. The problem of incomplete modalities increases the difficulty of understanding conversations accurately \cite{yang2018semi, xue2019deep, ma2021smil}.

	To this end, researchers propose various methods to deal with incomplete modalities. However, existing approaches are mainly designed for individual utterances or medical images rather than conversational data \cite{pham2019found, zhang2018multi}. For example, Pham et al. \cite{pham2019found} took modality-incomplete utterances as the input. They learned utterance-level representations via cyclic translations to ensure robustness to missing modalities. To diagnose Alzheimer's disease from incomplete medical images, Suo et al. \cite{suo2019metric} jointly optimized image imputation and metric learning. Liu et al. \cite{liu2021incomplete} predicted missing parts by distilling knowledge from complete data to incomplete data, resulting in better performance on medical images with incomplete modalities. Unlike individual utterances or medical images, conversations contain rich temporal and speaker information \cite{lian2019conversational, lian2021decn}. On the one hand, adjacent utterances are usually semantically related in a conversation. On the other hand, each speaker has their means of expression, generally consistent in a conversation. However, existing works usually fail to exploit them \cite{zhang2022deep, zhao2021missing}, thus limiting their performance in conversational data.

	In this paper, we propose a novel framework for incomplete multimodal learning in conversations called ``Graph Complete Network (GCNet)''. We aim to take full advantage of temporal and speaker information in conversations to deal with incomplete modalities. Fig. \ref{Figure1} shows the overall structure of our method. Specifically, we first randomly discard multimodal features to mimic real-world missing patterns \cite{zhang2022deep, chen2020hgmf}. To capture speaker and temporal dependencies in conversations, we propose two graph neural network-based modules, ``Speaker GNN (SGNN)'' and ``Temporal GNN (TGNN)''. These two modules share the same edges but different edge types. Finally, we jointly optimize classification and reconstruction in an end-to-end manner. To verify the effectiveness of our method, we conduct experiments on three benchmark conversational datasets. Through quantitative and qualitative analysis, we prove that our GCNet outperforms currently advanced approaches in both classification and imputation. The main contribution of this paper can be summarized as follows:

	\begin{itemize}
		\item Unlike existing works that mainly focus on medical images or individual utterances, we study the problem of incomplete modalities on conversational data, filling the gap of current works.
		
		\item We design a novel framework, GCNet, to deal with incomplete conversational data. This model leverages graph neural networks to capture temporal and speaker information in conversations.
		
		\item Experimental results on three benchmark datasets verify the effectiveness of our method. GCNet is superior to existing state-of-the-art approaches in incomplete multimodal learning.
		
	\end{itemize}

	The remainder of this paper is organized as follows: In Section \ref{sec2}, we briefly review some recent works on incomplete multimodal learning. In Section \ref{sec3}, we propose a novel framework for conversational data with incomplete modalities. In Section \ref{sec4}, we introduce experimental datasets and setup in detail. In Section \ref{sec5}, we conduct experiments to verify the effectiveness of our method. Finally, we conclude this paper and discuss future work in Section \ref{sec6}.

	\section{Related Works}
	\label{sec2}
	Learning from incomplete multimodal data is a fundamental research area in machine learning. A straightforward approach is to conduct data imputation and then utilize existing classification methods on the imputed data. In addition to imputation methods, there are also some strategies that can directly conduct learning without imputation.
	
	\subsection{Imputation Methods}
	Imputation methods attempt to estimate missing data from partially observed input. We review previous works and roughly divide them into three groups: zero/average imputation, low-rank imputation and DNN-based imputation.
	
	\textbf{Zero/average Imputation:} Padding missing modalities with zero vectors or average values are widely utilized for data imputation \cite{parthasarathy2020training, ma2021maximum, zhang2022deep}. For example, Parthasarathy et al. \cite{parthasarathy2020training} filled missing frames of videos with zero vectors. Zhang et al. \cite{zhang2022deep} padded missing modalities with average values based on the available samples within the same class. Zero/average imputation can achieve competitive performance in incomplete multimodal learning. However, since no supervision information is utilized, there is still a gap between filled values and original data, thus degrading the performance of downstream tasks.
	
	\textbf{Low-rank Imputation:} Complete multimodal data exhibits correlations between different modalities and leads to the low-rank data matrix. However, incomplete data breaks these correlations and increases tensor rank \cite{yang2017learning, liang2019learning}. To capture multimodal correlations, previous works project data into a common space by using low-rankness. These approaches are usually based on nuclear norm minimization, such as singular value thresholding (SVT) \cite{cai2010singular} and Soft-Impture \cite{mazumder2010spectral}. Besides nuclear norm, Fan et al. \cite{fan2017hyperspectral} also minimized tensor tubal rank to deal with various missing patterns. Furthermore, Liang et al. \cite{liang2019learning} combined the strength of non-linear functions to learn complex correlations in tensor rank minimization. However, these methods are usually computationally expensive for big data \cite{bishop2006pattern}.
	
	\textbf{DNN-based Imputation:} Due to the generative ability of DNNs, several DNN-based models have emerged to estimate missing data from partially observed input, e.g., autoencoder \cite{vincent2008extracting, tran2017missing}, GAN \cite{wang2018partial, cai2018deep}, VAE \cite{ivanov2018variational, du2018semi} and Transformer \cite{vaswani2017attention, yuan2021transformer}. Among these approaches, autoencoder and its variants are widely utilized due to their promising results in incomplete multimodal learning \cite{zhao2021missing}. For example, Duan et al. \cite{duan2014deep} leveraged autoencoders to impute missing data. To improve the modeling ability of autoencoders, Tran et al. \cite{tran2017missing} proposed the cascaded residual autoencoder (CRA). It combined a series of residual autoencoders \cite{he2016deep} into a cascaded architecture for data imputation. Furthermore, Zhao et al. \cite{zhao2021missing} incorporated CRA with cycle consistency loss for cross-modal imputation, which achieved superior performance over existing methods.

	\subsection{Non-imputation Methods}
	Existing non-imputation methods can be roughly divided into grouping strategies, correlation maximization and encoderless models.
	
	\textbf{Grouping Strategy:} Complete data is easier to deal with than incomplete data. The grouping strategy directly partitions incomplete data into multiple complete subgroups, and then feature learning is carried out independently for each subgroup \cite{yuan2012multi, xiang2013multi, li2018multi}. Despite its effectiveness, the number of subgroups grows exponentially with the number of modalities. Therefore, this strategy cannot work well for data with a large number of modalities or limited samples.

	\begin{figure*}[t]
		\centering
		\includegraphics[width=0.96\linewidth]{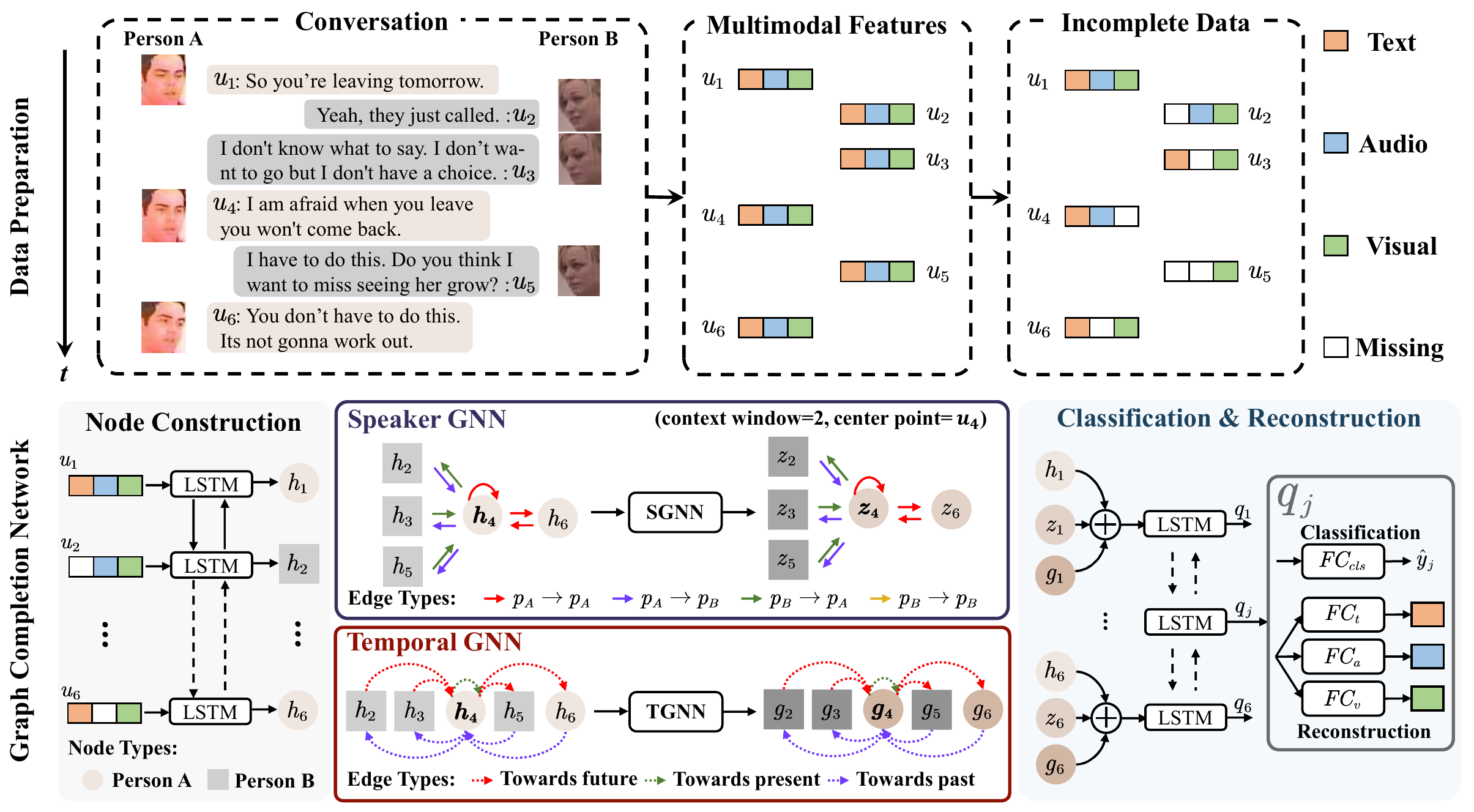}
		\caption{The overall structure of Graph Complete Network (GCNet) with the trimodal setting ($M=3$). This model focuses on the classification task based on incomplete conversational data.}
		\label{Figure1}
	\end{figure*}

	\textbf{Correlation Maximization:} To deal with the problem of incomplete data, an efficient approach is to maximize correlations between different modalities. In this way, we can constrain different modalities of the same sample to have related low-dimensional representations. Recently, several works based on correlation maximization have been proposed, including canonical correlation maximization \cite{hotelling1992relations, andrew2013deep}, HGR correlation maximization \cite{ma2021efficient}, mutual information maximization \cite{lin2021completer} and likelihood maximization \cite{ma2021maximum}. Among these approaches, canonical correlation and its variants are widely utilized due to their promising results in incomplete multimodal learning. For example, Hotelling et al. \cite{hotelling1992relations} proposed CCA that learned relationships between multi-modalities by linearly mapping them into a low-dimensional common space with maximal canonical correlations. Different from CCA that focused on linear mappings, Andrew et al. \cite{andrew2013deep} proposed DCCA that leveraged deep neural networks to learn more complex nonlinear combinations between multi-modalities. Wang et al. \cite{wang2015deep} further combined canonical correlations with reconstruction errors of autoencoders, trading off the structure information of each modality and the relationship between multi-modalities.

	\textbf{Encoderless Model:} Unlike previous works that rely on encoders, encoderless models can learn latent representations without encoders. They directly optimize latent representations to reconstruct modality-incomplete data regardless of missing patterns \cite{zadeh2019variational, zadeh2021relay}. Typically, Zhang et al. \cite{zhang2022deep} proposed CPM-Net, a robust encoderless model for incomplete multimodal learning. It combined the encoderless model with a clustering-like classification loss to learn well-structured features, which has validated its effectiveness on multimodal data with missing modalities.

	\section{Methodology}
	\label{sec3}
	In this paper, we focus on the classification task based on the conversational data with missing modalities. Each conversation contains multiple continuous utterances. For each utterance, we first extract multimodal features with random missing patterns. Then we propose a novel graph neural network-based framework, GCNet, to deal with incomplete conversational data. Fig. \ref{Figure1} shows the overall structure of our proposed method.

	\subsection{Data Preparation}
	Let us define a conversation $C=\{ \left(u_{i}, y_{i}\right)\}_{i=1}^{L}$, where $L$ is the number of utterances in the conversation, $u_{i}$ is the $i^{th}$ utterance in $C$ and $y_{i}$ is the true label of $u_{i}$. Here, $y_{i} \in \left\{1,2,\cdots, c\right\}$ and $c$ is the total number of labels. Each utterance $u_{i}$ is uttered by the speaker $p_{s(u_i)}$, where the function $s(\cdot)$ maps the index of utterance into its corresponding speaker. For each utterance $u_{i}$, we extract multimodal features $x_{i}=\{{x}_{i}^m\}_{m\in\{a, l, v\}}$. Here, $x_{i}^a \in \mathbb{R}^{d_a}$, $x_{i}^l \in \mathbb{R}^{d_l}$ and $x_{i}^v \in \mathbb{R}^{d_v}$ are the utterance-level features of acoustic, lexical and visual modalities, respectively. And $\{d_{m}\}_{m\in\{a, l, v\}}$ is the feature dimension of each modality.
	
\begin{figure}[t]
		\centering
		\includegraphics[width=0.88\linewidth]{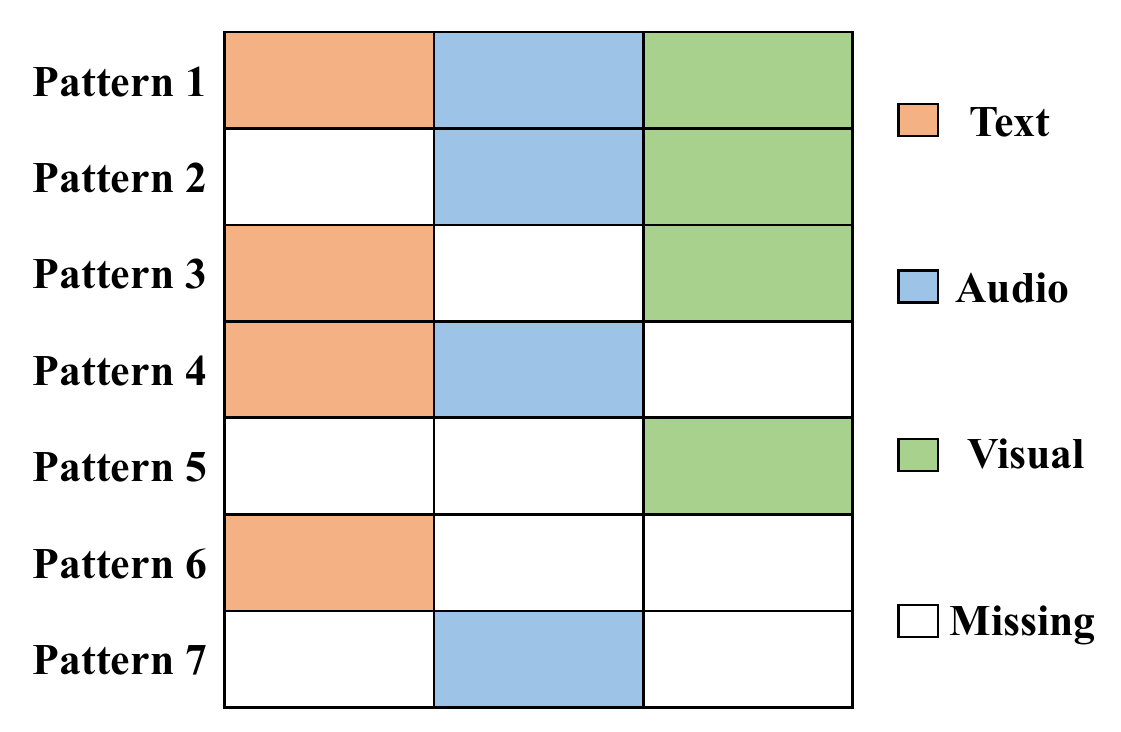}
		\caption{Seven missing patterns for a trimodal dataset.}
		\label{Figure2}
	\end{figure}

	To mimic real-world missing scenarios, we randomly discard some modalities by guaranteeing at least one modality is available for each sample, in line with previous works \cite{chen2020hgmf, zhang2022deep}. Therefore, an incomplete $M$-modal dataset has $\left(2^{M}-1\right)$ different missing patterns. Fig. \ref{Figure2} illustrates a trimodal ($M=3$) dataset with seven missing patterns. Suppose $\sigma_{i}$ is the missing pattern of $u_{i}$ and $\phi(\cdot)$ is a function that maps each missing pattern to its available modalities. The incomplete representation of $u_{i}$ is denoted as $\widetilde{x}_{i}=\{{\lambda}_{i}^m{x}_{i}^m\}_{m\in\{a, l, v\}}$, where ${\lambda}_{i}^m$ is defined as follows: 
	\begin{equation}
	\label{eq1}
	{\lambda}_{i}^m=\begin{cases}
	1, m\in\phi(\sigma_{i}) \\
	0, m\notin\phi(\sigma_{i}) \\
	\end{cases}
	\end{equation}
	
	\subsection{Graph Completion Network}
	\label{method-graphModel}
	Recent works have verified the importance of temporal and speaker information in conversations \cite{poria2017context, majumder2019dialoguernn}. To this end, we leverage this information to improve classification performance on incomplete conversational data. Our method consists of three key modules: Node Construction, Speaker \& Temporal GNN and Classification \& Reconstruction. 
	
	\subsubsection{Node Construction}
	We take the conversational data with missing modalities as the input. Each conversation contains multiple utterances, and we represent each utterance $u_i$ as a node $v_i$. To extract initial representations of $v_i$, we first concatenate incomplete multimodal features $\widetilde{x}_{i}=\{{\lambda}_{i}^m{x}_{i}^m\}_{m\in\{a, l, v\}}$, followed with bi-directional long short-term memory (Bi-LSTM) to capture contextual information. Bi-LSTM consists of gating mechanisms to control the flow of information, which is capable of modeling long-term contextual dependencies in both forward and backward directions:
	\begin{equation}
	{f}_i = Concat\left({\lambda}_{i}^a{x}_{i}^a, {\lambda}_{i}^l{x}_{i}^l, {\lambda}_{i}^v{x}_{i}^v\right),
	\end{equation}
	\begin{equation}
	H = \emph{BiLSTM}\left({F}, \theta_h \right),
	\end{equation}
	where ${F}=\{ {f}_i\}_{i=1}^{L} \in \mathbb{R}^{L \times (d_a+d_l+d_v)}$ and $H=\{ h_i\}_{i=1}^{L} \in \mathbb{R}^{L \times d}$ are the input and the output of Bi-LSTM, respectively. Here, $d$ is the feature dimension of output features. And $\theta_h$ is the trainable parameters. Finally, $H=\{ h_i\}_{i=1}^{L}$ is utilized as the initial node representations.

	\subsubsection{Speaker \& Temporal GNN}
	Speaker GNN (SGNN) and Temporal GNN (TGNN) are key modules for capturing speaker and temporal dependencies in conversations. In these modules, edges measure the importance of connections between nodes. Edge types define different aggregation approaches in node interactions \cite{yao2019graph, velivckovic2018graph}. SGNN and TGNN share the same edges but distinct edge types to capture different dependencies. In this section, we describe these two modules in detail.

	\textbf{Edges:} For each node, its interaction with the context nodes should be considered. If each node $v_i$ interacts with all context nodes $\{v_j\}_{j \in [1, L]}$, the designed graph will contain a large number of edges, making it difficult to store and optimize. Inspired by previous works that most utterances focus on their local context \cite{lian2021ctnet}, we fix the size of the context window to $w$, and each node $v_i$ only interacts with the nodes within the context window $\{v_j\}_{j \in \left[max(i-w, 1), min(i+w, L)\right]}$. In our implementation, we choose $w$ from $\{1, 2, 3, 4\}$. And we use $e_{ij}$ to represent the edge from $v_i$ to $v_j$.
	
	\textbf{Speaker Types:} SGNN leverages speaker types to capture speaker-sensitive dependencies in conversations. Specifically, we assign each edge $e_{ij}$ with a speaker type identifier $\alpha_{ij} \in \boldsymbol{\alpha}$. Here, $\boldsymbol{\alpha}$ denotes the set of speaker types, and $\left|\boldsymbol{\alpha}\right|$ is the number of $\boldsymbol{\alpha}$. For each edge $e_{ij}$, $\alpha_{ij}$ is set to $\left(p_{s(u_i)} \rightarrow p_{s(u_j)}\right)$, where $p_{s(u_i)}$ and $p_{s(u_j)}$ represent the speaker identity of $u_i$ and $u_j$, respectively. If there are $S$ distinct speakers in a conversation, there can be a maximum of $\left|\boldsymbol{\alpha}\right|=S^2$ distinct speaker types.

	\textbf{Temporal Types:} TGNN leverages temporal types to capture temporal-sensitive dependencies in conversations. Specifically, we assign each edge $e_{ij}$ with a temporal type identifier $\beta_{ij} \in \boldsymbol{\beta}$. Here, $\boldsymbol{\beta}$ denotes the set of temporal types, and $\left|\boldsymbol{\beta}\right|$ is the number of $\boldsymbol{\beta}$. Depending on the relative position of occurrence of $v_i$ and $v_j$ in a conversation, we determine the value of $\beta_{ij}$ to be either of \{past, present, future\}, resulting $\left|\boldsymbol{\beta}\right|=3$.

	\textbf{Graph Learning:} Recent works have verified the effectiveness of relation graph convolutional networks (R-GCN) in relation modeling problems (such as link prediction and entity classification) \cite{schlichtkrull2018modeling}. Inspired by its success, we leverage R-GCN to aggregate the neighborhood information in the graph. For SGNN and TGNN, we pass information via edges with parameters dependent on edge types. The calculation formula is shown as follows:
	\begin{equation}
	z_i=\sigma\left(\sum_{r \in \boldsymbol{\alpha}}\sum_{j \in {N}_i^r}{\frac{1}{\left|N_i^r\right|}}W_r^sh_j\right),
	\end{equation}
	\begin{equation}
	g_i=\sigma\left(\sum_{r \in \boldsymbol{\beta}}\sum_{j \in {N}_i^r}{\frac{1}{\left|N_i^r\right|}}W_r^th_j\right),
	\end{equation}
	where $z_i \in \mathbb{R}^h$ and $g_i \in \mathbb{R}^h$ are the outputs of SGNN and TGNN, respectively. Here, ${N}_i^r$ represents the set of neighboring indexes of the node $v_i$ under relation $r$, and $\left|{N}_i^r \right|$ is the number of ${N}_i^r$. $\sigma(\cdot)$ is the activation function, and we choose $ReLU(\cdot)$ is our implementation. $W_r^s$ and $W_r^t$ are the trainable parameters for SGNN and TGNN under relation $r$, respectively.

	\subsubsection{Classification \& Reconstruction}
	For each node $v_i$, we extract its initial representation $h_i$, a representation $z_i$ considering speaker information, and a representation $g_i$ considering temporal information. To aggregate these representations, we concatenate them together and then utilize Bi-LSTM for context-sensitive modeling:
	\begin{equation}
	\hat{h}_i=Concat\left(h_i, z_i, g_i\right),
	\end{equation}
	\begin{equation}
	Q = \emph{BiLSTM}\left({\hat{H}}, \theta_q\right),
	\end{equation}
	where $\hat{H}=\{\hat{h}_i\}_{i=1}^{L} \in \mathbb{R}^{L \times (3h)}$ and ${Q}=\{{q}_i\}_{i=1}^{L} \in \mathbb{R}^{L \times h}$ are the input and the output of Bi-LSTM, respectively. Here, $\theta_q$ is the trainable parameters. 
	
	\textbf{Classification}: To learn more discriminative features for conversation understanding, we feed the latent representations ${Q}=\{{q}_i\}_{i=1}^{L}$ into a fully-connected layer, followed by a softmax layer to estimate classification probabilities:
	\begin{equation}
	\hat{Y} = softmax(QW_c + b_c),
	\end{equation}
	where $\hat{Y}=\{\hat{y}_i\}_{i=1}^{L} \in \mathbb{R}^{L \times c}$ is the estimated probabilities, where $c$ is the number of discrete labels in the corpus. Here, $W_c \in \mathbb{R}^{d \times c}$ and $b_c \in \mathbb{R}^{c}$ are the trainable parameters.

	\textbf{Reconstruction}: Reconstructing complete data from the latent space can guide the model to learn the semantics of missing parts \cite{yuan2021transformer}. Therefore, we propose a modality-specific reconstruction module. For each modality $m \in \{a, l, v\}$, we perform a linear transformation mapping the extracted features into the input space:
	\begin{equation}
	\hat{X}^m = QW_m + b_m, m \in \{a, l, v\},
	\end{equation}
	where $\hat{X}^m=\{\hat{x}_i\}_{i=1}^{L} \in \mathbb{R}^{L \times d_m}$ is the estimated complete data. $W_m \in \mathbb{R}^{d \times d_m}$ and $b_c \in \mathbb{R}^{d_m}$ are the trainable parameters, where $d_m$ is the feature dimension for each modality.
	
	\subsubsection{Joint Optimization}
	To leverage both complete and incomplete data, we jointly optimize classification and imputation in an end-to-end manner. Therefore, our loss function consists of two parts: the classification loss $L_{cls}$ and the reconstruction loss $L_{rec}$.
	
	Minimizing the classification loss ensures learning more discriminative features, making the margins between different classes more explicit. During training, we choose the cross-entropy loss as the classification loss. The calculation formula is shown as follows:
	\begin{equation}
	L_{cls} = -\frac{1}{{L}}{\sum_{i=1}^{L}}y_{i}\log(\hat{y}_{i}),
	\end{equation}
	where ${y}_i \in \mathbb{R}^c$ is the true one-hot label. 
	
	To better estimate missing data from partially observed input, we calculate the reconstruction loss between the original and filled features at the missing positions:
	\begin{equation}
	\label{eq12}
	L_{rec} = \sum_{m \in \{a, l, v\}}{{\frac{1}{d_mL}\sum_{i=1}^{L}\lVert (1-\lambda_{i}^m) (\hat{x}_{i}^m-{x}_{i}^m) \rVert^2}},
	\end{equation}
	where $\lambda_{i}^m$ reveals available modalities for each utterance $u_i$, as defined in Eq. \ref{eq1}. Therefore, $(1-\lambda_{i}^m)$ is the mask revealing missing positions. 
	
	Finally, we combine these two loss functions into a joint objective function. This joint loss is utilized to optimize all trainable parameters in an end-to-end manner.
	\begin{equation}
	L = L_{cls} + L_{rec}
	\end{equation}
	
	This paper assumes that modality-complete data is available during training, consistent with previous works \cite{tran2017missing, zhao2021missing}. In practice, we first collect some modality-complete data using specific sensors. The inability to collect any such data rarely happens, which is left for our future work. Then, we can obtain a robust classifier for missing conditions by jointly optimizing the classification and reconstruction modules. During inference, we only need to leverage the pretrained classification module for conversation understanding. Therefore, although the raw information of missing modalities is unavailable in real-world scenarios, it does not affect our inference process.

	\section{Experimental Databases and Setup}
	\label{sec4}
	In this section, we first describe three benchmark conversational datasets in our experiments. Following that, we illustrate evaluation metrics, implementation details and multimodal features in detail. Finally, we introduce various currently advanced baselines in incomplete multimodal learning for comparison.
	
	\subsection{Corpus Description}
	\label{Sec4-1}
	To verify the effectiveness of GCNet, we conduct experiments on three benchmark conversational datasets, including IEMOCAP \cite{busso2008iemocap}, CMU-MOSI \cite{zadeh2016multimodal} and CMU-MOSEI \cite{zadeh2018multimodal}.

	\textbf{IEMOCAP} contains conversations where two actors perform improvised or scripted scenarios, especially chosen to evoke emotional expressions. These conversations are divided into five sessions. Each conversation is segmented into multiple utterances and each utterance is annotated with categorical labels. For a fair comparison, we follow two popular label process manners, resulting in four-class \cite{poria2017context, hazarika2018conversational} and six-class \cite{mai2020modality, majumder2019dialoguernn} datasets. Since no predefined data split manner is provided, we perform five-fold cross-validation using the leave-one-session-out strategy, in line with previous works \cite{lian2018speech, zhao2021combining}. We calculate the statistics of each session in Table \ref{Table1}.

	\textbf{CMU-MOSI} is a collection of movie review videos from online websites. To reflect sentiment intensity, each utterance is annotated with a sentiment score from -3 (extremely negative sentiment) to +3 (extremely positive sentiment).

	\textbf{CMU-MOSEI} is created by extending CMU-MOSI with more utterances and a greater variety of topics. Following the annotation manner in CMU-MOSI, each utterance is annotated with a sentiment score between [-3, 3]. Since train/validation/test splits are provided in CMU-MOSI and CMU-MOSEI, we utilize the official dataset split manner for a fair comparison. The statistics are shown in Table \ref{Table2}.

	\subsection{Evaluation Metrics}
	To evaluate GCNet against previous methods, we choose the following evaluation metrics for a fair comparison.
	
	\textbf{IEMOCAP} is labeled in categorical labels. Due to the natural imbalance across different classes \cite{busso2008iemocap, soldner2019box}, we use weighted average F1-score (WAF) as our evaluation metric, in line with previous works \cite{poria2017context, majumder2019dialoguernn}.
	
	\textbf{CMU-MOSI and CMU-MOSEI} are annotated in continuous sentiment scores between [-3, 3]. In this paper, we focus on the negative/positive classification task. Positive and negative classes are assigned for $<0$ and $>0$ scores, respectively. We also utilize WAF as our evaluation metric, in line with previous works \cite{hazarika2020misa, sun2020learning}.

	\begin{table}[t]
		\centering
		\renewcommand\tabcolsep{5.0pt}
		\renewcommand\arraystretch{1.20}
		\caption{Statistical information on IEMOCAP.}
		\label{Table1}
		\begin{tabular}{cc|c|c}
			\hline
			\multicolumn{2}{c|}{Dataset} & {\# conversations} & {\# utterances} \\
			\hline \hline
			
			\multicolumn{1}{c|}{\multirow{5}{*}{\begin{tabular}[c]{@{}c@{}}IEMOCAP \\ (four-class)\end{tabular}}}
			&\multicolumn{1}{|c|}{Session1} &28 &1085 \\
			&\multicolumn{1}{|c|}{Session2} &30 &1023 \\
			&\multicolumn{1}{|c|}{Session3} &32 &1151 \\
			&\multicolumn{1}{|c|}{Session4} &30 &1031 \\
			&\multicolumn{1}{|c|}{Session5} &31 &1241 \\
			\hline
			
			\multicolumn{1}{c|}{\multirow{5}{*}{\begin{tabular}[c]{@{}c@{}}IEMOCAP \\ (six-class)\end{tabular}}}
			&\multicolumn{1}{|c|}{Session1} &28 &1373 \\
			&\multicolumn{1}{|c|}{Session2} &30 &1356 \\
			&\multicolumn{1}{|c|}{Session3} &32 &1569 \\
			&\multicolumn{1}{|c|}{Session4} &30 &1512 \\
			&\multicolumn{1}{|c|}{Session5} &31 &1623 \\
			\hline

		\end{tabular}
	\end{table}

	\begin{table}[t]
		\centering
		\renewcommand\tabcolsep{5.0pt}
		\renewcommand\arraystretch{1.20}
		\caption{Statistical information on CMU-MOSI and CMU-MOSEI.}
		\label{Table2}
		\begin{tabular}{cc|c|c|c|c|c|c}
			\hline
			\multicolumn{2}{c|}{\multirow{2}{*}{Dataset}} & \multicolumn{3}{c|}{\# conversations} & \multicolumn{3}{c}{\# utterances} \\
			&& train & val				& test	& train & val 	& test 	\\ \hline \hline
			
			\multicolumn{2}{c|}{CMU-MOSI} & 52 & 10 	 & 31 & 1284 & 229	& 686	\\ 
			\hline
			
			\multicolumn{2}{c|}{CMU-MOSEI} 	& 2249 & 300 & 676 & 16326 & 1871 & 4659 \\ 
			\hline
			
		\end{tabular}
	\end{table}

	\subsection{Implementation Details}
	We investigate the performance of different methods on multimodal datasets with varying missing rates. The missing rate is defined as $\eta = 1-\frac{\sum_{i=1}^{L}{m_i}}{L \times M}$, where $m_i$ indicates the number of available modalities for the $i^{th}$ sample. Here, $L$ denotes the total number of samples and $M$ represents the total number of modalities. For each sample with $M$ modalities, we randomly select the missing modalities with the probability $\eta$. We also guarantee that at least one modality is available for each sample, resulting $m_i \geqslant 1$ and $\eta \leqslant \frac{M-1}{M}$. Therefore, if the number of modalities is set to $M=3$, we choose the missing rate $\eta$ from $[0.0, 0.1, \cdots, 0.7]$, where $0.7$ is an approximation of $\frac{M-1}{M}$ with the same effect. We keep the same missing rate during training, validation and testing periods, in line with previous works \cite{zhang2022deep, yuan2021transformer}.

	Since train/validation/test splits are provided in the CMU-MOSI and CMU-MOSEI datasets, we select the best model on the validation set and report its performance on the test set. For the IEMOCAP dataset, we turn all parameters with five-fold cross-validation. There are two user-specific parameters in our GCNet, i.e., the dimension of latent representations $h$ and the context window size $w$. We select $h$ from $\{50, 100, 200\}$ and $w$ from $\{1, 2, 3, 4\}$ for all datasets. During training, we use the Adam optimization scheme with a learning rate of 0.001 and a weight decay of 0.00001. To alleviate the over-fitting problem, Dropout \cite{srivastava2014dropout} is also utilized with a rate of $p=0.5$. Since our GCNet takes conversation-level features as the input, we pad conversations of the same mini-batch to have the same number of utterances. Bit masking is also implemented to eliminate the effect of padded parts. To evaluate the performance of different methods, we run each experiment ten times and report the average values on the test set.

	\subsection{Multimodal Feature Extraction}
	\label{ses:multi-feat}
	 For each utterance, we extract multimodal features from acoustic, lexical and visual modalities. The multimodal feature extraction process is described as follows:
	
	\textbf{Acoustic Modality:} Pre-trained wav2vec \cite{schneider2019wav2vec} is used as the acoustic feature extractor. This model is a multi-layer convolutional neural network trained on a large amount of unlabeled data. Inspired by its success in many downstream tasks \cite{fan2020exploring}, we leverage the pre-trained \emph{wav2vec-large}\footnote{\emph{https://github.com/pytorch/fairseq/tree/main/examples/wav2vec}} to extract 512-dimensional acoustics features for each utterance.
	
	\textbf{Lexical Modality:} Pre-trained DeBERTa \cite{he2020deberta} is exploited as the lexical feature extractor. This model improves BERT \cite{devlin2018bert} and RoBERTa \cite{liu2019roberta} using the disentangled attention mechanism and the enhanced mask decoder. Inspired by its success in both natural language understand and natural language generation \cite{he2020deberta}, we leverage the pre-trained \emph{DeBERTa-large}\footnote{\emph{https://huggingface.co/microsoft/deberta-large}} to encode word sequences into 1024-dimensional lexical features for each utterance.
	
	\textbf{Visual Modality:} Pre-trained MA-Net \cite{zhao2021learning} is utilized as the visual feature extractor. This model leverages global multi-scale and local attention to address occlusions and non-frontal poses. Inspired by its success in facial emotion recognition, we use the MTCNN face detection algorithm to extract aligned faces \cite{zhang2016joint} and then employ the pre-trained MA-Net\footnote{\emph{https://github.com/zengqunzhao/MA-Net}} for facial feature extraction. Finally, we compress frame-level facial features into 1024-dimensional utterance-level features by average encoding.

	\subsection{Baselines}
	To evaluate the performance of our proposed GCNet, we implement the following state-of-the-art incomplete multimodal learning methods as the baselines.
	
	\textbf{CCA} \cite{hotelling1992relations} is a strong benchmark model. To deal with the problem of incomplete data, it learns relationships across different modalities by linearly mapping them into low-dimensional common space with maximum correlations.
	
	\textbf{DCCA} \cite{andrew2013deep} is an extension of CCA. CCA focuses on linear combinations of different modalities. Differently, DCCA leverages deep neural networks to learn more complex nonlinear combinations between multi-modalities.
	
	\textbf{DCCAE} \cite{wang2015deep} is another extension of CCA. It employs autoencoders to learn hidden features for each modality, and then optimizes both reconstruction errors of autoencoders and canonical correlations. This objective function trades off the structure information of each modality and the inherent association between multiple modalities.
	
	\textbf{AE} \cite{bengio2007greedy} is widely utilized in incomplete multimodal learning \cite{duan2014deep, wong2014imputing}. It leverages autoencoders to impute missing data from partially observed input. Followed with previous works \cite{zhao2021missing, yuan2021transformer}, we jointly optimize the reconstruction loss of autoencoders and the classification loss of downstream tasks in our implementation.
	
	\textbf{CRA} \cite{tran2017missing} is an extension of AE. It combines a series of residual autoencoders into a cascaded architecture for data imputation. In our implementation, we jointly optimize imputation and downstream tasks end-to-end.
	
	\textbf{MMIN} \cite{zhao2021missing} is another extension of AE. It combines CRA with cycle consistency learning to predict latent representations of missing modalities. This model is a strong benchmark model, which outperforms other approaches under varying missing conditions.
	
	\textbf{CPM-Net} \cite{zhang2022deep} jointly considers completeness and structure to learn discriminative latent representations. As for completeness, it designs an encoderless model that projects all samples into a common space regardless of missing patterns. As for structuring, it learns well-structured features by equipping with a clustering-like classification loss.
	
	\section{Results and Discussion}
	\label{sec5}
	In this paper, we propose a novel framework, GCNet, for incomplete multimodal learning in conversations. To verify the effectiveness of our method, we first conduct comparative experiments with currently advanced approaches, investigating the classification and imputation performance under different missing rates. Then, we show the importance of incomplete data in multimodal learning. Next, we reveal the necessity of each component in GCNet, including SGNN for speaker-sensitive modeling and TGNN for temporal-sensitive modeling. After that, we reveal the impact of different hyper-parameters and study the convergence property. To qualitatively analyze the improvement of our proposed method, we further visualize latent representations and conduct case studies.

	\begin{table*}[htbp]
		\centering
		\renewcommand\tabcolsep{8pt}
		\renewcommand\arraystretch{1.20}
		\caption{Comparison of classification performance with different missing rates. We report WAF scores(\%), and higher WAF indicates better performance. The best performance is highlighted in bold, and the second-highest result is labeled by $^\dag$. The row with $\Delta_{\emph{SOTA}}$ means the improvements or reductions of GCNet compared to existing state-of-the-art systems.}
		\label{Table3}
		\begin{tabular}{c|c|c|c|c|c|c|c|c|c|c}
			\hline
			\multirow{2}{*}{Dataset}													&
			\multirow{2}{*}{Method}														&
			\multicolumn{8}{c}{\begin{tabular}[c]{@{}c@{}}Missing Rate\end{tabular}}	\\ \cline{3-11}
			& & 0.0	&0.1 &0.2 &0.3 &0.4 &0.5 & 0.6 & 0.7 & Average \\
			\hline \hline

			\multirow{8}{*}{\begin{tabular}[c]{@{}c@{}}IEMOCAP \\ (four-class)\end{tabular}}	 
			&CCA \cite{hotelling1992relations} 	& 64.52	& 65.19	& 62.60	& 59.35 & 55.25	& 51.38	& 45.73	& 30.61 & 54.33	\\
			&DCCA \cite{andrew2013deep}	& 60.03	& 57.25	& 51.74	& 42.53	& 36.54	& 34.82	& 33.65	& 41.09 & 44.71	\\
			&DCCAE \cite{wang2015deep} 	& 63.42	& 61.66	& 57.67	& 54.95	& 51.08	& 45.71	& 39.07	& 41.42 & 51.87	\\
			&CPM-Net \cite{zhang2022deep} & 58.00	& 55.29	& 53.65	& 52.52	& 51.01	& 49.09 & 47.38 & 44.76 & 51.46 \\
			&AE \cite{bengio2007greedy}	& 74.82	& 71.36	& 67.40	& 62.02	& 57.24	& 50.56	& 43.04 & 39.86 & 58.29	\\
			&CRA \cite{tran2017missing}	& 76.26$^\dag$	& 71.28	& 67.34	& 62.24	& 57.04	& 49.86	& 43.22	& 38.56 & 58.23	\\
			&MMIN \cite{zhao2021missing} & 74.94 & 71.84$^\dag$ & 69.36$^\dag$ & 66.34$^\dag$ & 63.30$^\dag$ & 60.54$^\dag$ & 57.52$^\dag$ & 55.44$^\dag$ & 64.91$^\dag$ \\
			&\textbf{GCNet} & \textbf{78.36} & \textbf{77.48}	& \textbf{77.34}	& \textbf{76.22}	& \textbf{75.14}	& \textbf{73.80}	& \textbf{71.88}	& \textbf{71.38} & \textbf{75.20}	\\
			&$\Delta_{\emph{SOTA}}$		
			&\textcolor[rgb]{0.0,0.6,0.0}{$\uparrow$2.10}
			&\textcolor[rgb]{0.0,0.6,0.0}{$\uparrow$5.64}
			&\textcolor[rgb]{0.0,0.6,0.0}{$\uparrow$7.98}
			&\textcolor[rgb]{0.0,0.6,0.0}{$\uparrow$9.88}
			&\textcolor[rgb]{0.0,0.6,0.0}{$\uparrow$11.84}
			&\textcolor[rgb]{0.0,0.6,0.0}{$\uparrow$13.26}
			&\textcolor[rgb]{0.0,0.6,0.0}{$\uparrow$14.36}
			&\textcolor[rgb]{0.0,0.6,0.0}{$\uparrow$15.94}
			&\textcolor[rgb]{0.0,0.6,0.0}{$\uparrow$10.29}
			\\
			\hline
			\hline

			\multirow{8}{*}{\begin{tabular}[c]{@{}c@{}}IEMOCAP \\ (six-class)\end{tabular}}	 
			&CCA \cite{hotelling1992relations} & 43.04 & 46.06 & 43.86 & 41.66 & 37.13 & 34.94 & 32.06 & 21.80 & 37.57 \\
			&DCCA \cite{andrew2013deep}	& 42.18 	& 39.15	& 34.47	& 27.65	& 23.69		& 22.86		& 22.71		& 27.38	& 30.01	\\
			&DCCAE \cite{wang2015deep}	& 46.19		& 43.77		& 41.28		& 37.98		& 34.58		& 30.02		& 26.78		& 27.66	& 36.03	\\
			&CPM-Net \cite{zhang2022deep}	& 41.05		& 37.33		& 36.22		& 35.73		& 35.11		& 33.64		& 32.26		& 31.25	& 35.32	\\
			&AE \cite{bengio2007greedy}	 & 56.76		& 52.82		& 48.66		&42.26		& 35.18		& 29.12		& 25.08		& 23.18	& 39.13	\\
			&CRA \cite{tran2017missing}				& \textbf{58.68}		& 53.50		& 49.76		& 45.88		& 39.94		& 32.88		& 28.08		& 26.16	& 41.86 \\
			&MMIN \cite{zhao2021missing} 			& 56.96		& 53.94$^\dag$		& 51.46$^\dag$		& 48.42$^\dag$		& 45.60$^\dag$		& 42.82$^\dag$		& 40.18$^\dag$		& 37.84$^\dag$ & 47.15$^\dag$	\\
			&\textbf{GCNet} 		& 58.64$^\dag$		& \textbf{58.50}		& \textbf{57.64}		& \textbf{57.08}		& \textbf{56.12}	& \textbf{54.40}		& \textbf{53.60}		& \textbf{53.46} & \textbf{56.18}	\\
			&$\Delta_{\emph{SOTA}}$				&\textcolor[rgb]{0.5,0.5,0.5}{$\downarrow$0.04} 
			&\textcolor[rgb]{0.0,0.6,0.0}{$\uparrow$4.56}
			&\textcolor[rgb]{0.0,0.6,0.0}{$\uparrow$6.18}
			&\textcolor[rgb]{0.0,0.6,0.0}{$\uparrow$8.66}
			&\textcolor[rgb]{0.0,0.6,0.0}{$\uparrow$10.52}
			&\textcolor[rgb]{0.0,0.6,0.0}{$\uparrow$11.58}
			&\textcolor[rgb]{0.0,0.6,0.0}{$\uparrow$13.42}
			&\textcolor[rgb]{0.0,0.6,0.0}{$\uparrow$15.62}
			&\textcolor[rgb]{0.0,0.6,0.0}{$\uparrow$9.03}\\
			\hline
			\hline

			\multirow{8}{*}{\begin{tabular}[c]{@{}c@{}} CMU-MOSI \end{tabular}}	
			&CCA \cite{hotelling1992relations} & 74.74 & 71.60 & 68.84 & 65.71 & 62.13 & 60.37 & 59.92 & 53.26 & 64.57 \\		
			&DCCA \cite{andrew2013deep}	 & 67.60 & 65.67 & 61.46 & 58.81 & 54.43 & 52.77 & 50.13 & 53.29 & 58.02	\\
			&DCCAE \cite{wang2015deep}	 & 66.76 & 64.21 & 62.61 & 59.23 & 59.61 & 53.56 & 51.87 & 53.49 &	58.92\\
			&CPM-Net \cite{zhang2022deep} & 71.90 & 68.91 & 71.12 & 70.59 & 67.95 & 65.88 & 64.02 & 61.79$^\dag$ & 67.77 \\
			&AE	\cite{bengio2007greedy}	 & \textbf{85.86} & 82.27 & 78.43 & 74.00 & 69.83 & 66.62 & 60.22 & 55.64 & 71.61 \\		
			&CRA \cite{tran2017missing}		& 85.68$^\dag$ & \textbf{82.61} & 78.53$^\dag$ & 75.12$^\dag$ & 70.20$^\dag$ & 67.40 & 62.40 & 59.40 & 72.67\\
			&MMIN \cite{zhao2021missing} 		& 85.20 & 81.91 & 78.22 & 74.60 & 70.14 & 67.72$^\dag$ & 64.04$^\dag$ & 61.53 & 72.92$^\dag$\\		
			&\textbf{GCNet} & 85.01 & {82.54}$^\dag$ & \textbf{80.17} & \textbf{78.54} & \textbf{76.48} & \textbf{73.45} & \textbf{69.46} & \textbf{68.35} & \textbf{76.75} \\
			&$\Delta_{\emph{SOTA}}$		
			&\textcolor[rgb]{0.5,0.5,0.5}{$\downarrow$0.85} 
			&\textcolor[rgb]{0.5,0.5,0.5}{$\downarrow$0.07}
			&\textcolor[rgb]{0.0,0.6,0.0}{$\uparrow$1.64}
			&\textcolor[rgb]{0.0,0.6,0.0}{$\uparrow$3.42}
			&\textcolor[rgb]{0.0,0.6,0.0}{$\uparrow$6.28}
			&\textcolor[rgb]{0.0,0.6,0.0}{$\uparrow$5.73}
			&\textcolor[rgb]{0.0,0.6,0.0}{$\uparrow$5.42}
			&\textcolor[rgb]{0.0,0.6,0.0}{$\uparrow$6.56}
			&\textcolor[rgb]{0.0,0.6,0.0}{$\uparrow$3.83}\\
			\hline
			\hline

			\multirow{8}{*}{\begin{tabular}[c]{@{}c@{}} CMU-MOSEI \end{tabular}}	
			&CCA \cite{hotelling1992relations}	& 81.23	& 79.06	& 78.09	& 76.25	& 74.62	& 73.22	& 70.57	& 56.29 & 73.67	\\
			&DCCA \cite{andrew2013deep}	& 74.06	& 70.11	& 64.69	& 61.31	& 61.11	& 60.47	& 58.27	& 59.76 & 63.72 \\
			&DCCAE \cite{wang2015deep} & 75.70 & 74.67 & 73.16 & 71.82 & 70.26 & 67.86 & 64.15 & 62.75 & 70.05 \\
			&CPM-Net \cite{zhang2022deep} & 78.47 & 74.79 & 74.48 & 73.81 & 72.39 & 70.43 & 68.73 & 67.07 & 72.52 \\
			&AE	\cite{bengio2007greedy}	& 86.66$^\dag$ & 84.37$^\dag$ & 82.58$^\dag$ & 80.57$^\dag$ & 78.80$^\dag$ & 76.43$^\dag$ & 74.26$^\dag$ & 72.81$^\dag$ & 79.56$^\dag$\\
			&CRA \cite{tran2017missing}	& 86.48	& 84.19	& 82.25	& 80.12	& 78.55	& 75.85	& 74.07	& 72.46	& 79.25 \\
			&MMIN \cite{zhao2021missing} & 85.78 & 83.77 & 81.85 & 79.77 & 77.63 & 75.36 & 72.95 & 71.18 & 78.54\\
			&\textbf{GCNet} & \textbf{87.12} & \textbf{86.50} & \textbf{85.50} & \textbf{84.53} & \textbf{83.55} & \textbf{82.44} & \textbf{80.27} & \textbf{80.20} & \textbf{83.76} \\
			&$\Delta_{\emph{SOTA}}$		
			&\textcolor[rgb]{0.0,0.6,0.0}{$\uparrow$0.46} 
			&\textcolor[rgb]{0.0,0.6,0.0}{$\uparrow$2.13}
			&\textcolor[rgb]{0.0,0.6,0.0}{$\uparrow$2.92}
			&\textcolor[rgb]{0.0,0.6,0.0}{$\uparrow$3.96}
			&\textcolor[rgb]{0.0,0.6,0.0}{$\uparrow$4.75}
			&\textcolor[rgb]{0.0,0.6,0.0}{$\uparrow$6.01}
			&\textcolor[rgb]{0.0,0.6,0.0}{$\uparrow$6.01}
			&\textcolor[rgb]{0.0,0.6,0.0}{$\uparrow$7.39}
			&\textcolor[rgb]{0.0,0.6,0.0}{$\uparrow$4.20} \\
			\hline

		\end{tabular}
	\end{table*}

	\begin{figure*}[t]
		\begin{center}
			\subfigure[IEMOCAP(four-class)]{
				\label{Figure3-1}
				\centering
				\includegraphics[width=0.388\linewidth]{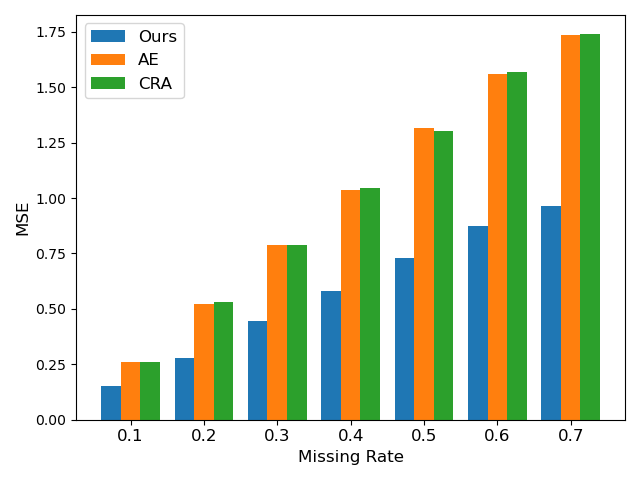}
			}
			\subfigure[IEMOCAP(six-class)]{
				\label{Figure3-2}
				\centering
				\includegraphics[width=0.388\linewidth]{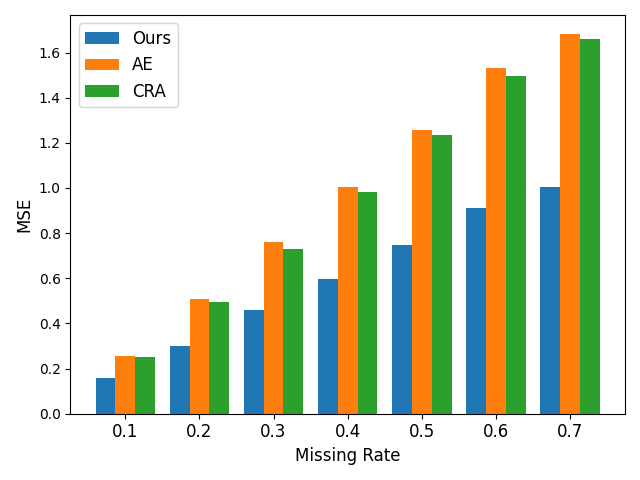}
			} \\
			\subfigure[CMU-MOSI]{
				\label{Figure3-3}
				\centering
				\includegraphics[width=0.388\linewidth]{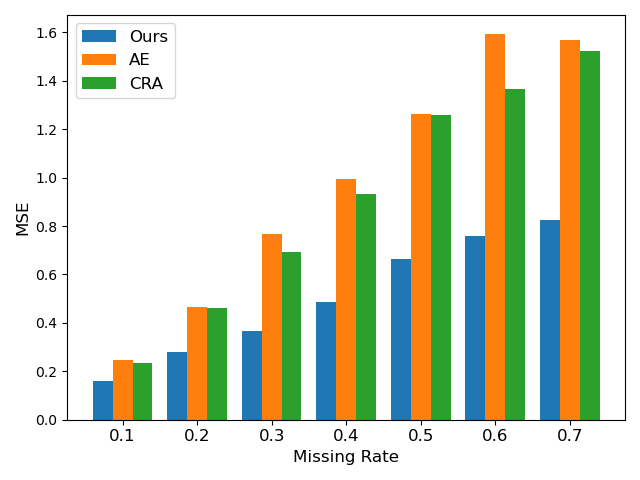}
			}
			\subfigure[CMU-MOSEI]{
				\label{Figure3-4}
				\centering
				\includegraphics[width=0.388\linewidth]{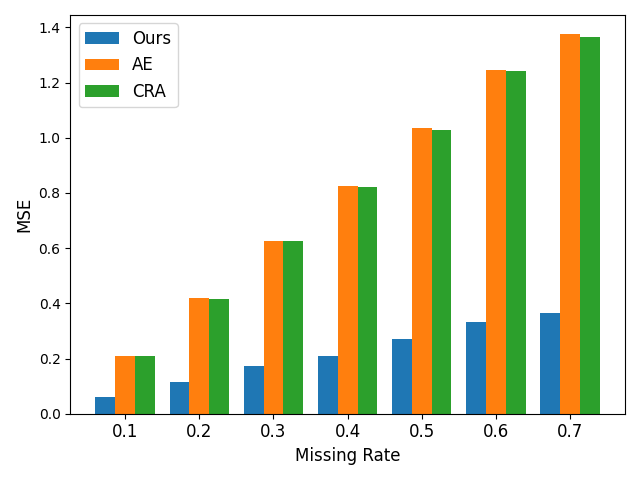}
			}
		\end{center}
		\caption{Comparison of imputation performance with different missing rates. Lower MSE indicates better imputation performance.}
		\label{Figure3}
	\end{figure*} 
	
	\subsection{Classification Performance}
	Table \ref{Table3} presents the classification performance of different approaches under varying missing rates. From these experimental results, we have the following observations:
	
	1. On average, our method achieves the best performance on all datasets. For the IEMOCAP(four-class) dataset, GCNet succeeds over currently advanced approaches by 10.29\%. For the IEMOCAP(six-class) dataset, GCNet achieves the new state-of-the-art record 56.18\%, which shows an absolute improvement of 9.03\%. We also observe a similar phenomenon on the CMU-MOSI and CMU-MOSEI datasets. These quantitative results verify the effectiveness of our method in incomplete multimodal learning. We argue that these baselines do not explicitly model temporal and speaker dependencies in conversations, which are essential for conversation understanding. Differently, the proposed method leverages graph neural networks to capture these dependencies, resulting in better classification performance.

	2. Experimental results in Table \ref{Table3} show that the performance degradation of our method is much smaller than that of baselines as the missing rate increases. Taking the results on the IEMOCAP(four-class) dataset as an example, as the missing rate increases from 0.0 to 0.7, the performance of baselines declines 13.24\%$\sim$37.70\% while our GCNet only declines 6.98\%. More notably, the performance gap between GCNet and baselines becomes more significant as the missing rate increases. Taking the results on the IEMOCAP(four-class) dataset as an example, the performance gap increases from 2.10\% to 15.94\% as the missing rate increases from 0.0 to 0.7. These results verify that GCNet improves classification performance, especially in severely missing cases.

	3. Experimental results in Table \ref{Table3} demonstrate that our method also exhibits competitive performance on complete multimodal data ($\eta=0.0$). For the IEMOCAP(four-class) dataset, our GCNet shows an absolute improvement of 2.10\% over currently advanced approaches. For the CMU-MOSEI dataset, our proposed method succeeds over state-of-the-art baselines by 0.46\%. Although our method achieves slightly worse performance than baselines on the CMU-MOSI and IEMOCAP(six-class) datasets, the difference is not significant. These results validate the effectiveness of our method on modality-complete data.

	\subsection{Imputation Performance}
	GCNet leverages graph neural networks to impute missing data from partially observed input. To verify the effectiveness of our method, we compare the imputation performance of GCNet with two currently advanced imputation methods, \textbf{AE} \cite{bengio2007greedy} and \textbf{CRA} \cite{tran2017missing}. To evaluate the imputation performance of different methods, we calculate the mean square error (MSE) between original features and estimated features at the missing position.

	Fig. \ref{Figure3} presents the imputation results of different methods under varying missing rates. Experimental results demonstrate that our GCNet consistently outperforms all baselines under all missing rates on all datasets. Speaker and temporal dependencies are crucial in data imputation. Firstly, adjacent utterances in a conversation are usually semantically related. Secondly, each speaker has their means of expression, which is generally consistent in a conversation. But most baselines ignore these dependencies, thus degrading their imputation performance. Differently, our GCNet makes full use of speaker and temporal information via graph neural networks and achieves better imputation performance. These results verify the importance of speaker and temporal information in data imputation and the effectiveness of our method in incomplete multimodal learning.

	Meanwhile, experimental results in Fig. \ref{Figure3} demonstrate that the imputation performance declines as the missing rate increases. It is reasonable because higher missing rates can result in fewer observed data, increasing the difficulty of data imputation. Meanwhile, the performance gap between GCNet and baselines becomes more significant as the missing rate increases. These results validate that our model is more robust to missing data than baselines.

	\begin{figure*}[t]
		\begin{center}
			\subfigure[IEMOCAP(four-class)]{
				\label{Figure4-1}
				\centering
				\includegraphics[width=0.388\linewidth]{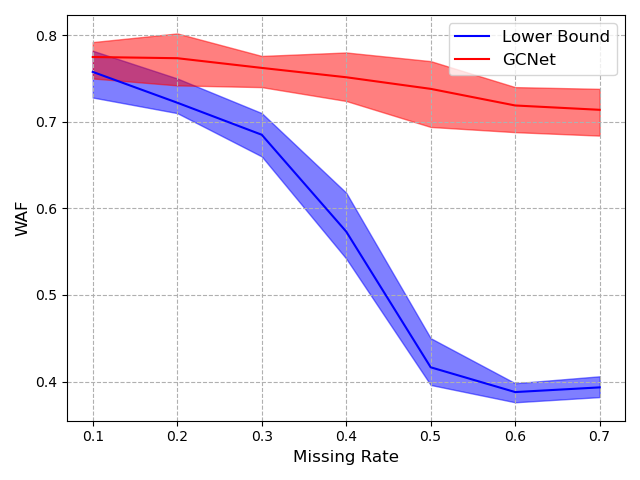}
			}
			\subfigure[IEMOCAP(six-class)]{
				\label{Figure4-2}
				\centering
				\includegraphics[width=0.388\linewidth]{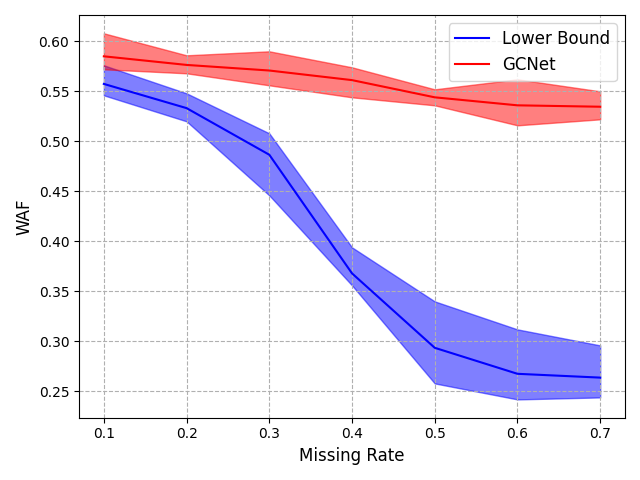}
			} \\
			\subfigure[CMU-MOSI]{
				\label{Figure4-3}
				\centering
				\includegraphics[width=0.388\linewidth]{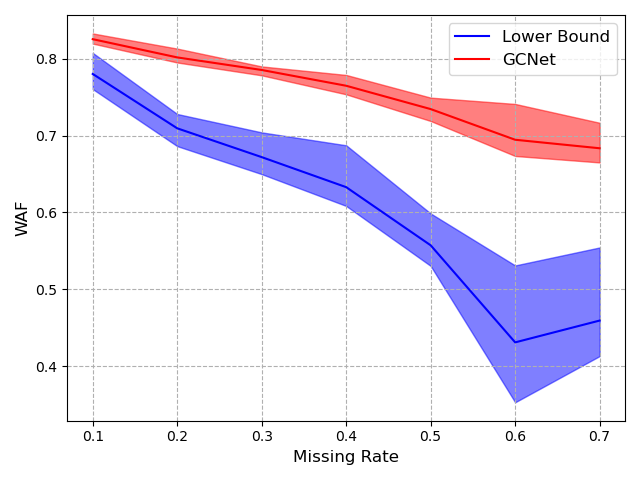}
			}
			\subfigure[CMU-MOSEI]{
				\label{Figure4-4}
				\centering
				\includegraphics[width=0.388\linewidth]{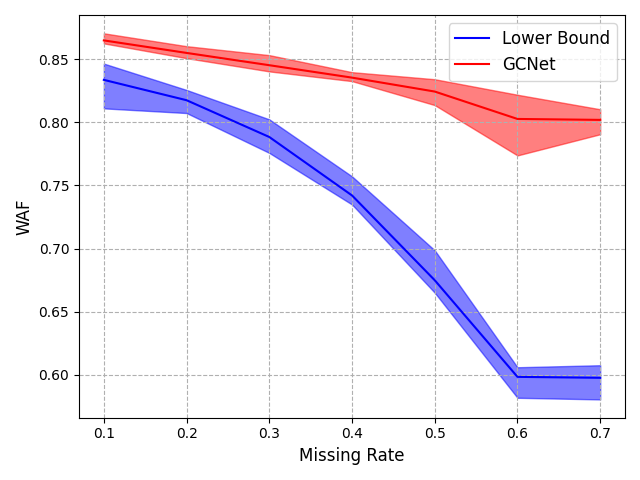}
			}
		\end{center}
		\caption{Classification performance comparison of GCNet and Lower Bound under different missing rates.}
		\label{Figure4}
	\end{figure*}

	\subsection{Importance of Incomplete Data}
	Besides modality-complete data, GCNet also makes full use of modality-incomplete data during training. To investigate the importance of modality-incomplete data, we implement one comparison system. In Fig. \ref{Figure4}, we compare the performance of different methods under varying missing rates.
	\begin{itemize}
		\item \textbf{GCNet}: It is our proposed method that leverages both modality-complete and modality-incomplete data in the learning process.
		
		\item \textbf{Lower Bound}: It comes from GCNet, but abandons modality-incomplete data and only focuses on modality-complete data. It is one of the most straightforward strategies to handle modality missing and is often regarded as the lower bound \cite{ma2021maximum}.
	\end{itemize}
	
	As shown in Fig. \ref{Figure4}, GCNet consistently outperforms the comparison system with all missing rates on all datasets. It is reasonable because the comparison system does not use extra information in modality-incomplete data, resulting in a loss of available information. 
	
	Meanwhile, the performance gap between GCNet and the comparison system becomes more significant as the missing rate increases. The reason lies in that the comparison system will delete a large number of training samples, especially in severe missing cases. It is challenging to learn discriminative classifiers with limited data. Differently, our GCNet incorporates both modality-complete data and modality-incomplete data in the learning process. With more available training samples, we can use the complementary information in these two types of data and learn more discriminative representations for classification.

	\subsection{Role of SGNN and TGNN}
	In GCNet, we capture speaker and temporal dependencies via SGNN and TGNN. To verify the necessity of these components, we implement three comparison systems. Experimental results are listed in Table \ref{Table4}.
	\begin{itemize}
		\item \textbf{GCNet}: It is our method that captures speaker and temporal dependencies via SGNN and TGNN.
		
		\item \textbf{GCNet-T}: It comes from GCNet but omits the SGNN component. Therefore, this model ignores speaker information in conversations.
		
		\item \textbf{GCNet-S}: It comes from GCNet but omits the TGNN component. Therefore, this model ignores temporal information in conversations.
		
		\item \textbf{GCNet-ST}: It comes from DialogueGCN \cite{ghosal2019dialoguegcn}, a currently advanced graphic model for conversation understanding. Unlike our approach that captures speaker and temporal dependencies through two separate graphs, this comparison system captures ``speaker-temporal'' dependencies simultaneously in one graph. This coupling strategy will increase relation types that need to be optimized.
	\end{itemize}
	
	Experimental results in Table \ref{Table4} demonstrate that GCNet exhibits performance improvement over \textbf{GCNet-S} in most cases. The difference between these two models lies in whether TGNN is utilized for temporal-sensitive modeling. Our method can leverage temporal information to learn more discriminative features, achieving better classification performance. These results verify the importance of temporal information in incomplete multimodal learning and the effectiveness of our method in temporal-sensitive modeling.
	
	Meanwhile, our GCNet outperforms \textbf{GCNet-T} on all missing rates. Taking the results on the IEMOCAP(six-class) dataset as an example, GCNet shows an absolute improvement of 0.31\%$\sim$2.35\% over \textbf{GCNet-T}. Compared with this comparison system, we further exploit SGNN to capture speaker information in conversations. Since each speaker has its means of expression, we can well reconstruct conversational data with the help of speaker information and achieve better classification performance. 
	
	From Table \ref{Table4}, we also observe that GCNet outperforms \textbf{GCNet-ST} in most cases. Although we can easily collect conversational data from social media platforms, the annotation process still requires a lot of manual effort, resulting in insufficient amounts of labeled data. Compared with GCNet, \textbf{GCNet-ST} needs to optimize more relation types with limited labeled data. It increases the difficulty of model optimization, and each relation type is hard to be fully learned. Our GCNet captures speaker and temporal dependencies through two separate models, which reduces the number of relation types, alleviates the difficulty of model optimization, and achieves better performance.

	\begin{table}[t]
		\centering
		\renewcommand\tabcolsep{12pt}
		\renewcommand\arraystretch{1.20}
		\caption{Ablation results for SGNN and TGNN on the IEMOCAP dataset. We report the classification performance of different models under varying missing rates $\eta$. The bold font represents the best performance.}
		\label{Table4}
		\begin{tabular}{c|c|c|c}
			\hline
			{\begin{tabular}[c]{@{}c@{}} $\eta$ \end{tabular}} & {Model} & {\begin{tabular}[c]{@{}c@{}}IEMOCAP \\ (four-class)\end{tabular}} & {\begin{tabular}[c]{@{}c@{}}IEMOCAP \\ (six-class)\end{tabular}} 
			\\ \cline{3-4}
			\hline 
			\hline
			0.0 & GCNet 	& \textbf{78.36}	& \textbf{58.64}	\\ 
			0.0	& GCNet-T	& 76.90 			& 57.43 			\\ 
			0.0	& GCNet-S	& 77.65 			& 58.03				\\ 
			0.0	& GCNet-ST	& 77.86				& 58.60 			\\
			\hline
			0.1 & GCNet 	& \textbf{77.48}	& \textbf{58.50}	\\ 
			0.1	& GCNet-T	& 75.60 			& 57.06 			\\ 
			0.1	& GCNet-S	& 76.80 			& 57.37 			\\ 
			0.1	& GCNet-ST	& 77.04				& 57.80 			\\
			\hline
			0.2 & GCNet 	& \textbf{77.34}	& \textbf{57.64} 	\\ 
			0.2	& GCNet-T	& 74.55				& 56.06 			\\ 
			0.2	& GCNet-S	& 75.80				& 56.94 			\\ 
			0.2	& GCNet-ST	& 76.70 			& 56.64 			\\
			\hline
			0.3 & GCNet 	& \textbf{76.22} 	& \textbf{57.08} 	\\ 
			0.3	& GCNet-T	& 74.05 			& 55.14				\\ 
			0.3	& GCNet-S	& 75.35 			& 55.54				\\ 
			0.3	& GCNet-ST	& 75.44				& 55.84 			\\
			\hline
			0.4 & GCNet 	& 75.14 			& \textbf{56.12} 	\\ 
			0.4	& GCNet-T	& 72.05				& 55.06 			\\ 
			0.4	& GCNet-S	& 74.25				& 55.20				\\ 
			0.4	& GCNet-ST	& \textbf{75.20}	& 55.40				\\
			\hline
			0.5 & GCNet 	& \textbf{73.80} 	& 54.40				\\ 
			0.5	& GCNet-T	& 72.10				& 54.09				\\ 
			0.5	& GCNet-S	& 72.85				& 54.14				\\ 
			0.5	& GCNet-ST	& 73.68				& \textbf{54.54}	\\
			\hline
			0.6 & GCNet 	& \textbf{71.88} 	& 53.60				\\ 
			0.6	& GCNet-T	& 71.40				& 52.40 			\\ 
			0.6	& GCNet-S	& 71.60				& \textbf{53.74}	\\ 
			0.6	& GCNet-ST	& 71.80 			& 52.94				\\
			\hline
			0.7 & GCNet 	& \textbf{71.38} 	& \textbf{53.46} 	\\ 
			0.7	& GCNet-T	& 71.20 			& 51.11 			\\ 
			0.7	& GCNet-S	& 71.30 			& 51.69 			\\ 
			0.7	& GCNet-ST	& 71.00 			& 50.36				\\
			\hline
		\end{tabular}
	\end{table}

	\begin{figure}[t]
		\centering
		\includegraphics[width=0.6\linewidth, trim=88 0 16 0]{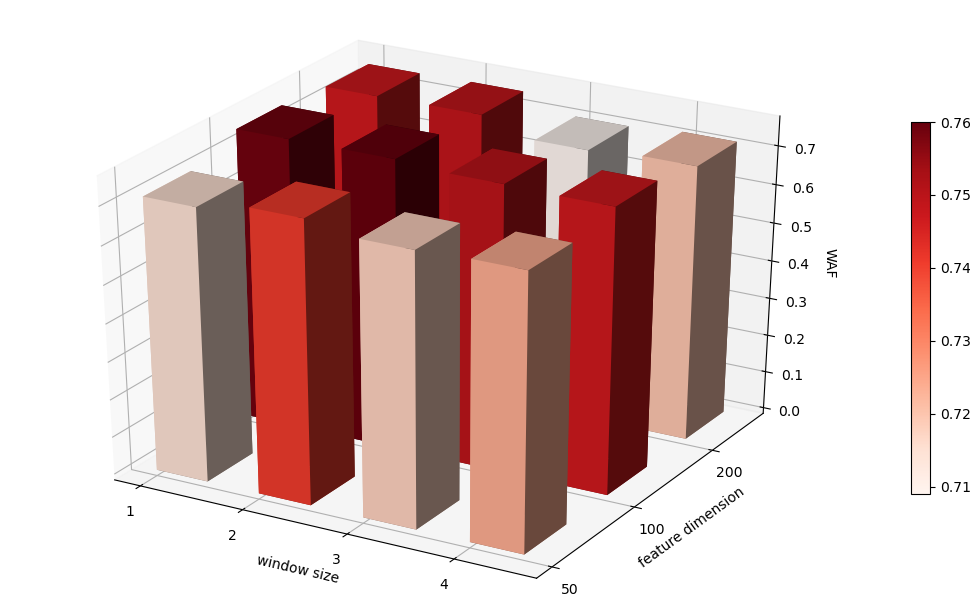}
		\caption{Parameter tuning on the IEMOCAP(four-class) dataset with the missing rate $\eta=0.3$.}
		\label{Figure5}
	\end{figure}

	\begin{figure}[t]
		\begin{center}
			\subfigure[classification loss]{
				\label{Figure6-1}
				\centering
				\includegraphics[width=0.472\linewidth, trim=15 0 15 0]{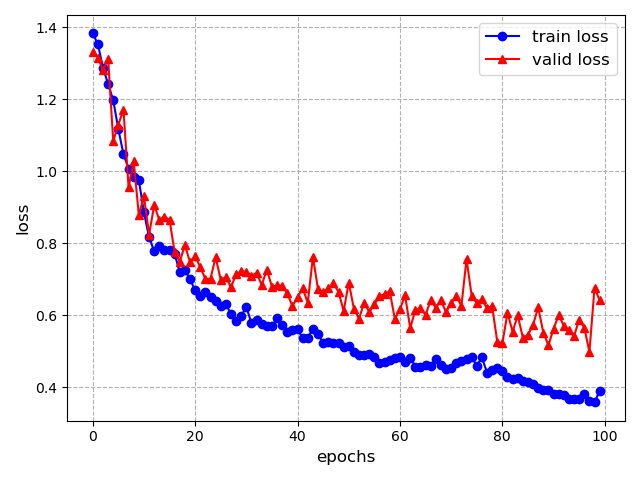}
			} 
			\subfigure[imputation loss]{
				\label{Figure6-2}
				\centering
				\includegraphics[width=0.472\linewidth, trim=15 0 15 0]{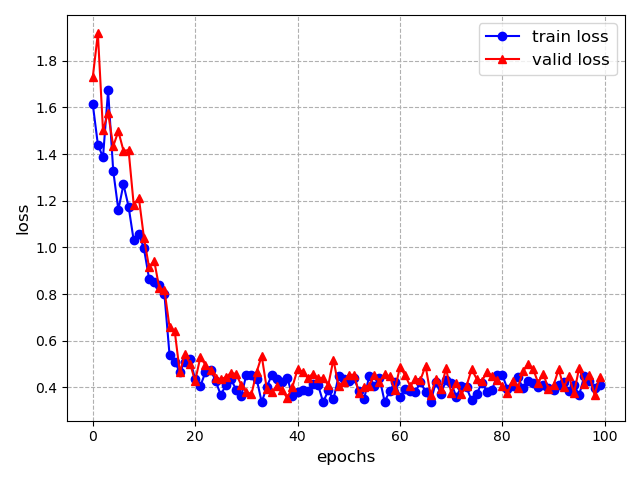}
			} 
		\end{center}
		\caption{Trends in loss functions on the IEMOCAP(four-class) dataset with the missing rate $\eta=0.3$.}
		\label{Figure6}
	\end{figure} 
	
	\begin{figure*}[t]
		\begin{center}
			\subfigure[CCA]{
				\label{Figure7-1}
				\centering
				\includegraphics[width=0.235\linewidth, trim=42 42 42 42] {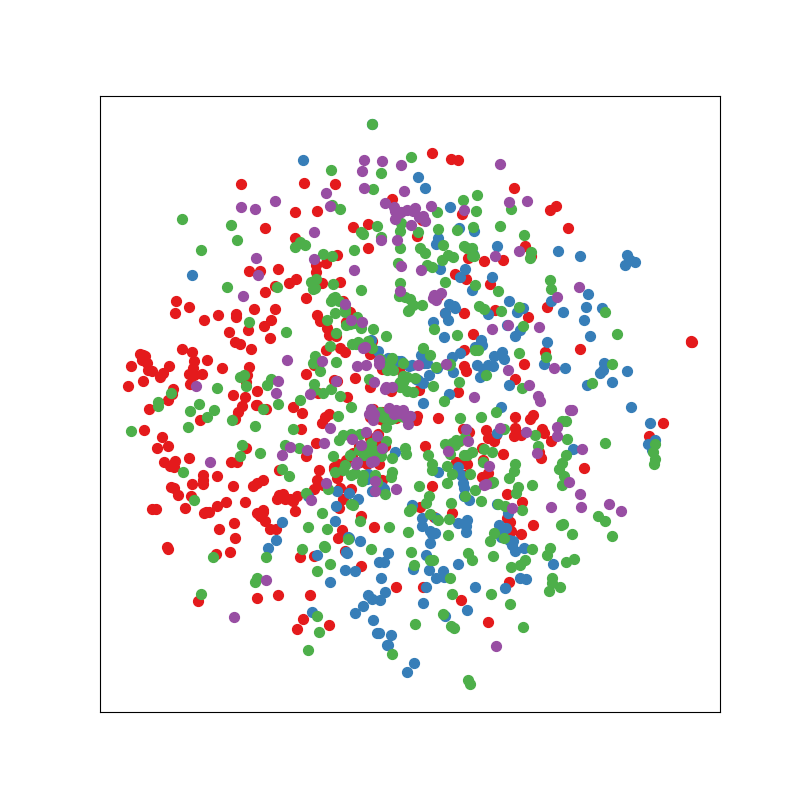}
			}
			\subfigure[DCCA]{
				\label{Figure7-2}
				\centering
				\includegraphics[width=0.235\linewidth, trim=42 42 42 42] {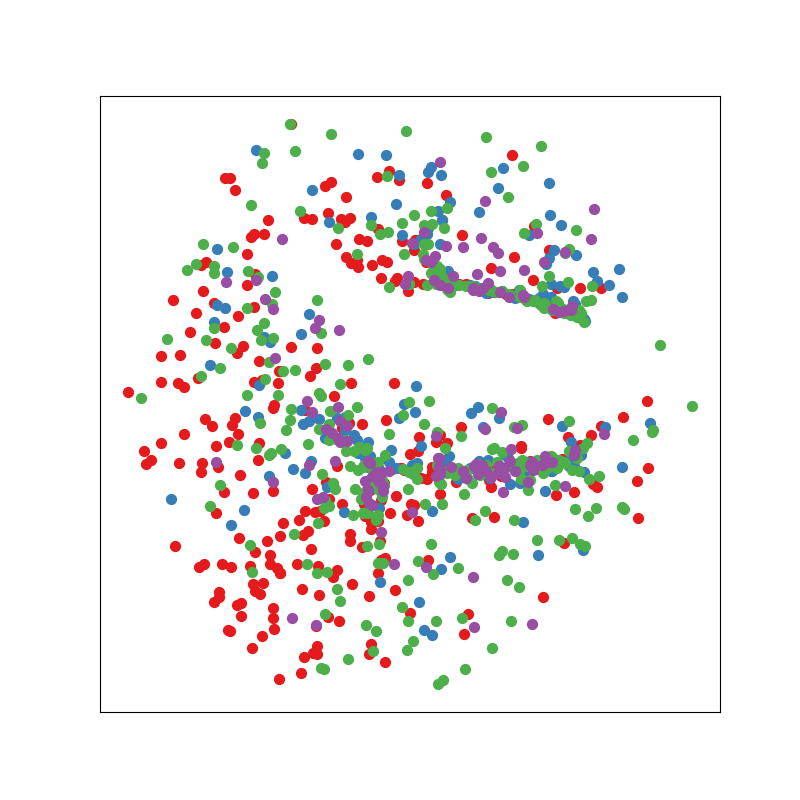}
			}
			\subfigure[DCCAE]{
				\label{Figure7-3}
				\centering
				\includegraphics[width=0.235\linewidth, trim=42 42 42 42] {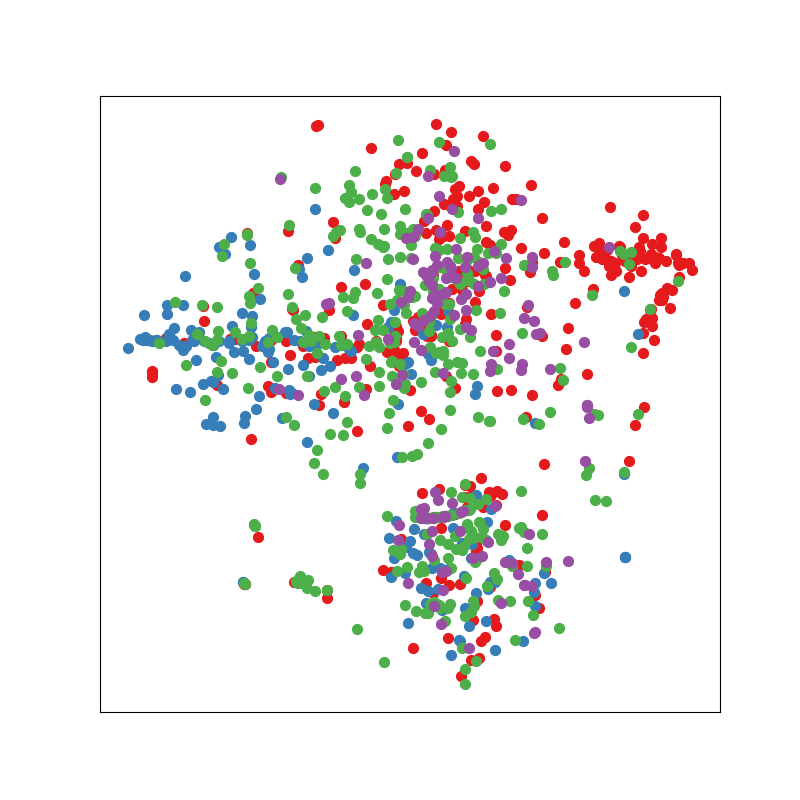}
			}
			\subfigure[CPM-Net]{
				\label{Figure7-4}
				\centering
				\includegraphics[width=0.235\linewidth, trim=42 42 42 42] {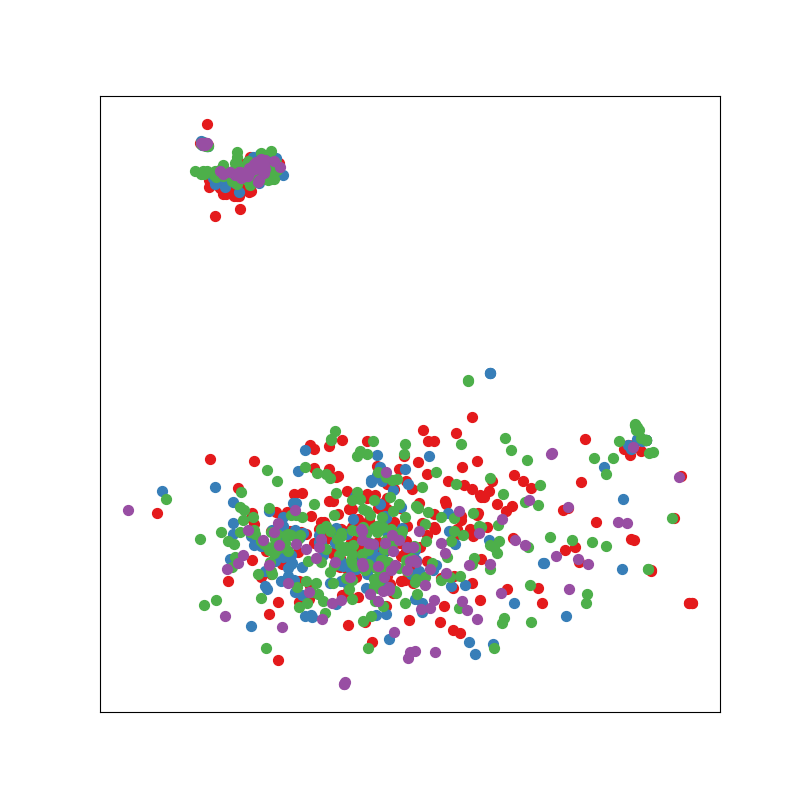}
			}
			\subfigure[AE]{
				\label{Figure7-5}
				\centering
				\includegraphics[width=0.235\linewidth, trim=42 42 42 42] {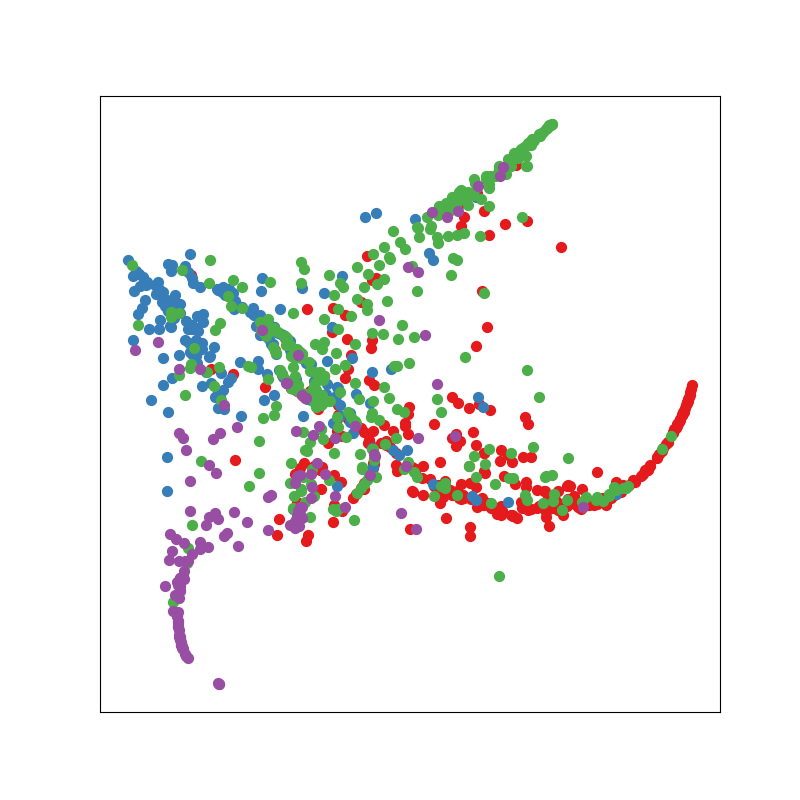}
			}
			\subfigure[CRA]{
				\label{Figure7-6}
				\centering
				\includegraphics[width=0.235\linewidth, trim=42 42 42 42] {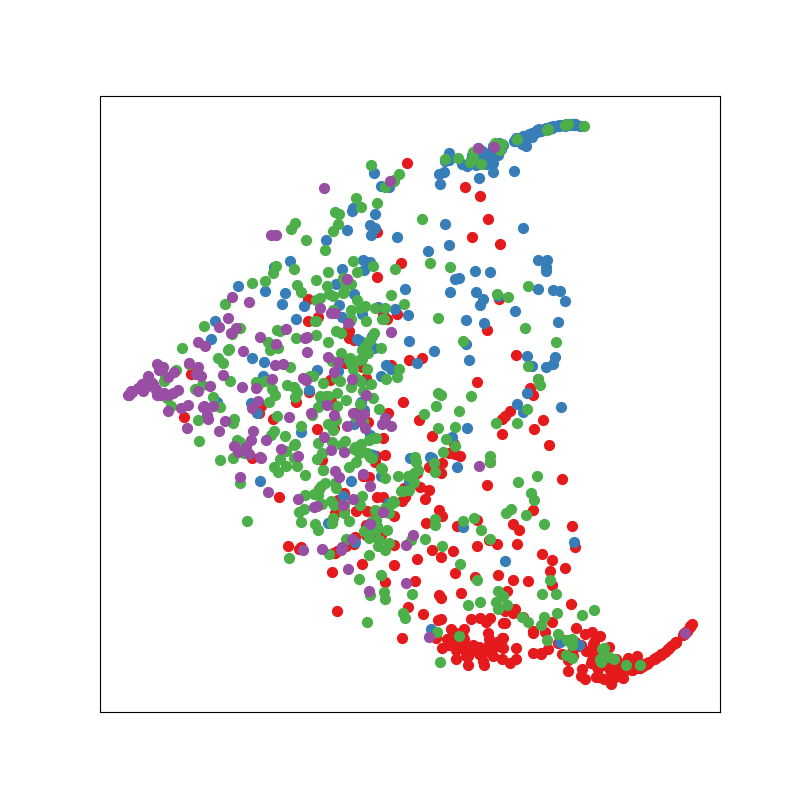}
			}
			\subfigure[MMIN]{
				\label{Figure7-7}
				\centering
				\includegraphics[width=0.235\linewidth, trim=42 42 42 42] {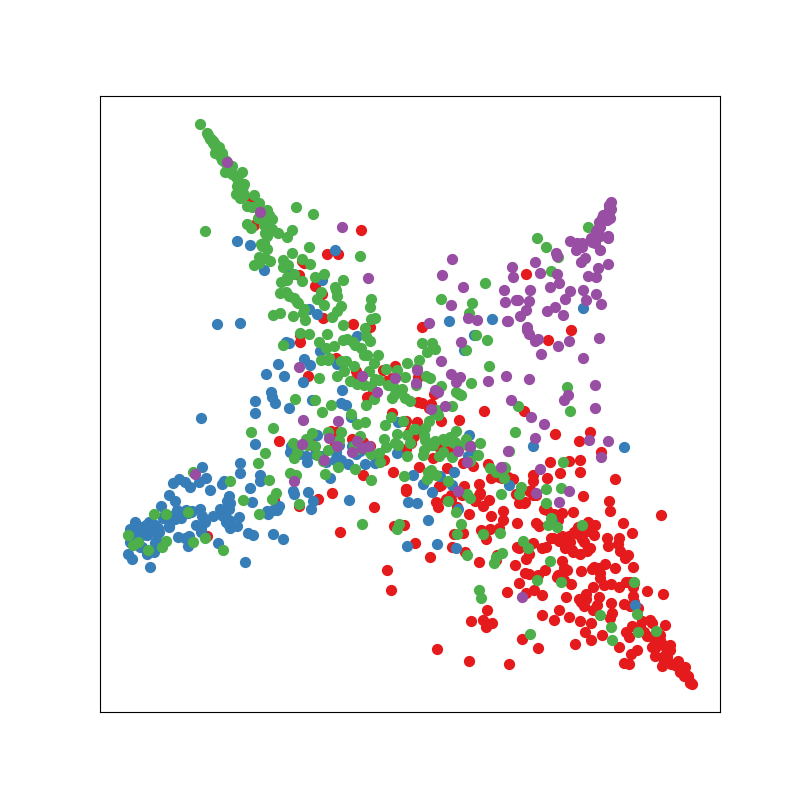}
			}
			\subfigure[GCNet]{
				\label{Figure7-8}
				\centering
				\includegraphics[width=0.235\linewidth, trim=42 42 42 42] {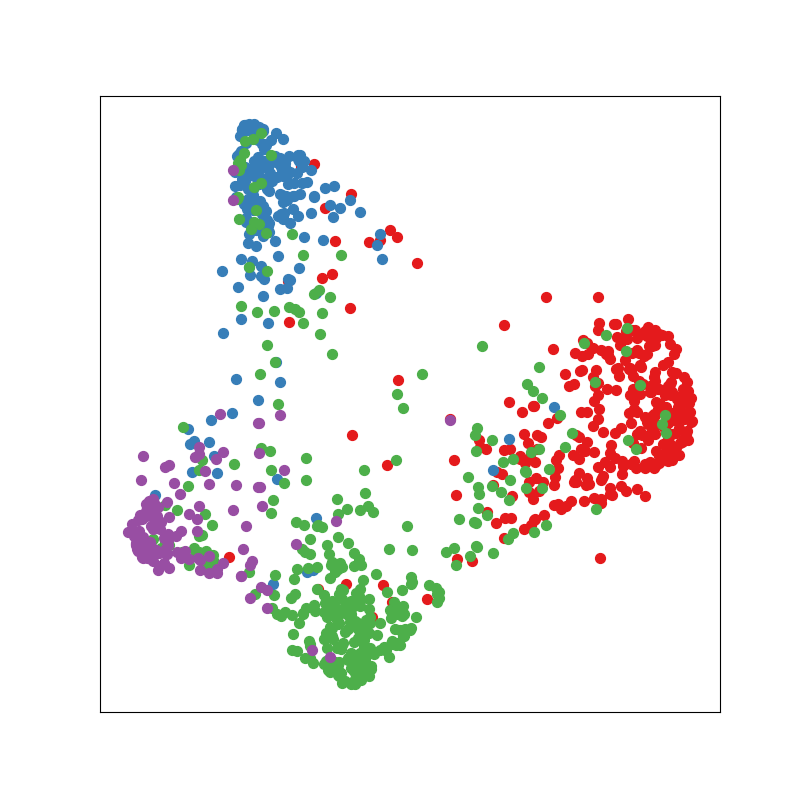}
			}
		\end{center}
		\caption{Visualization the representations of different methods on the IEMOCAP(four-class) test set with the missing rate $\eta=0.3$. In these figures, we use red, blue, green and purple to represent happiness, sadness, neutral and anger, respectively.}
		\label{Figure7}
	\end{figure*}

	\begin{figure*}[t]
		\begin{center}
			\subfigure[10 epoch]{
				\label{Figure8-1}
				\centering
				\includegraphics[width=0.235\linewidth, trim=42 42 42 42] {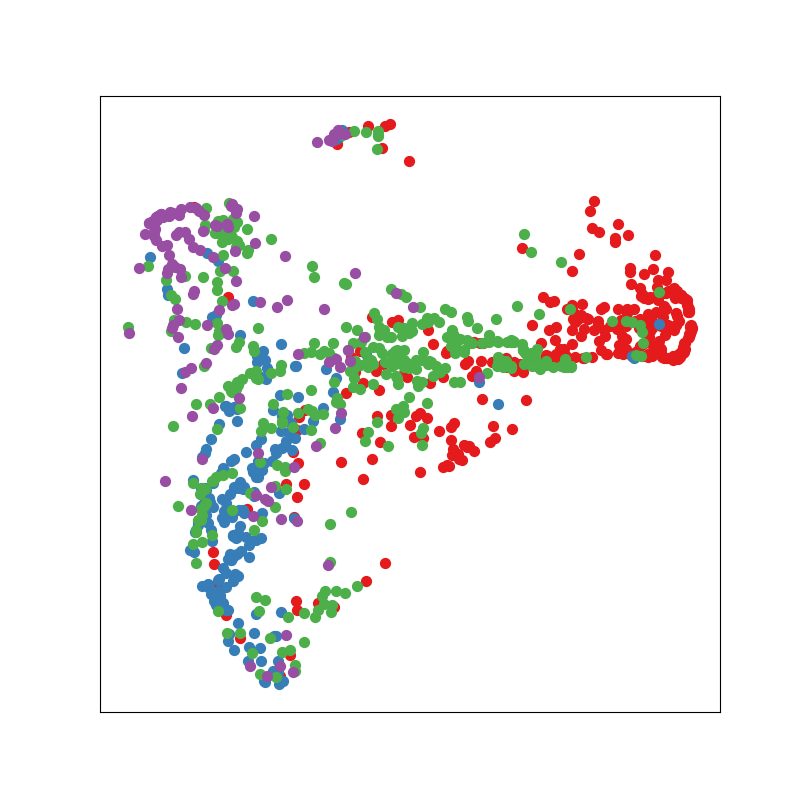}
			}
			\subfigure[20 epoch]{
				\label{Figure8-2}
				\centering
				\includegraphics[width=0.235\linewidth, trim=42 42 42 42] {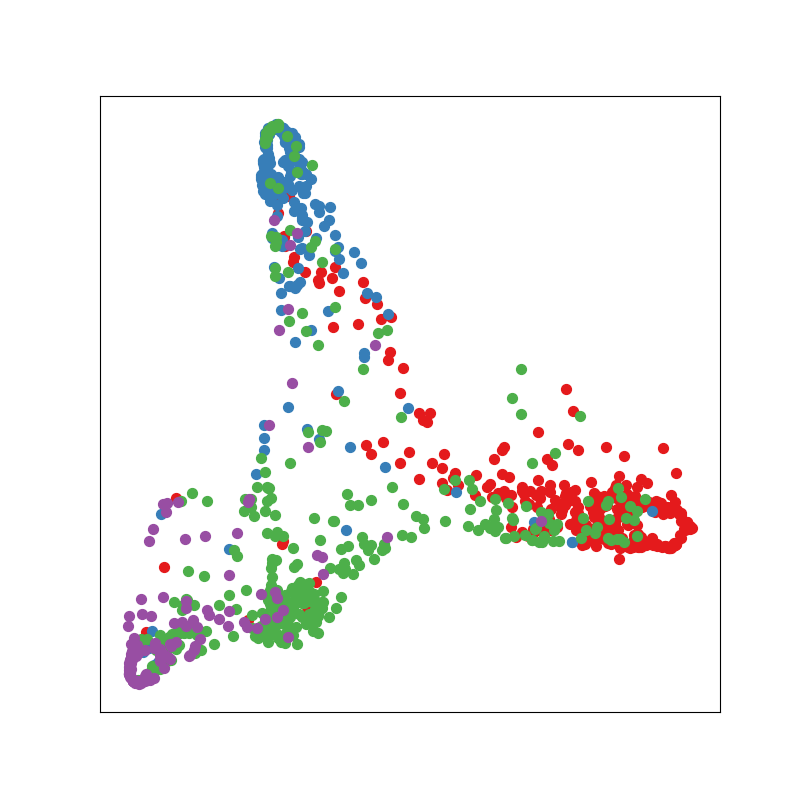}
			}
			\subfigure[50 epoch]{
				\label{Figure8-3}
				\centering
				\includegraphics[width=0.235\linewidth, trim=42 42 42 42] {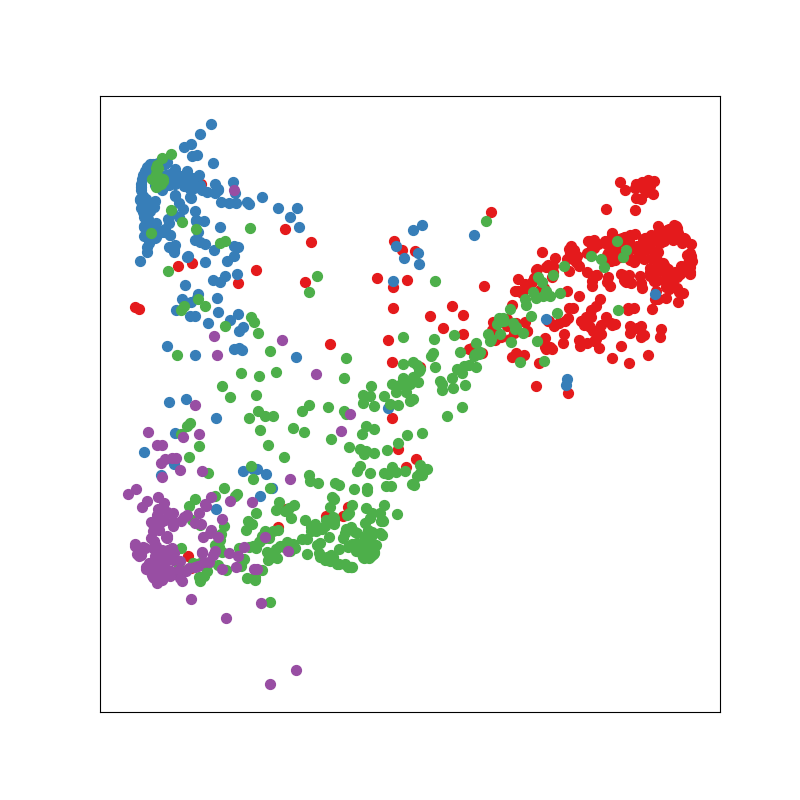}
			}
			\subfigure[78 epoch]{
				\label{Figure8-4}
				\centering
				\includegraphics[width=0.235\linewidth, trim=42 42 42 42] {image/graphlstm_best}
			}
		\end{center}
		\caption{T-SNE visualization results on the IEMOCAP(four-class) test set with increasing training iterations (the missing rate $\eta=0.3$). In these figures, we use red, blue, green and purple to represent happiness, sadness, neutral and anger, respectively.}
		\label{Figure8}
	\end{figure*}

	\subsection{Parameter Tuning}
	Our GCNet contains two user-specified parameters: the context window size $w$ and the dimension of latent representations $h$. To evaluate the influence of these parameters, we visualize the parameter tuning process on the IEMOCAP(four-class) dataset with the missing rate $\eta=0.3$. We choose $w$ from $\{1, 2, 3, 4\}$ and $h$ from $\{50, 100, 200\}$. Experimental results are presented in Fig. \ref{Figure5}.
	
	As $w$ increases, the classification performance improves first and then degrades. The reasons are probably twofold. Firstly, as the window size increases, the designed graph will contain a large number of edges, which increases the difficulty of model optimization. Secondly, most utterances focus on their local context \cite{lian2021ctnet}. A larger window size will include more irrelevant contextual information, which may affect the prediction results of the target utterance. Therefore, unlike previous works \cite{hu2021mmgcn} that exploit fully connected graphs for conversation understanding, we fix the context window size to limit the number of edges.

	Meanwhile, as $h$ gradually increases, the classification performance improves first and then degrades. When the feature dimension increases, our GCNet will contain more trainable parameters, increasing the risk of over-fitting. Therefore, a good choice of parameters can remarkably improve the performance of conversation understanding.

	\begin{figure*}[t]
		\centering
		\includegraphics[width=\linewidth, trim=0 10 0 0]{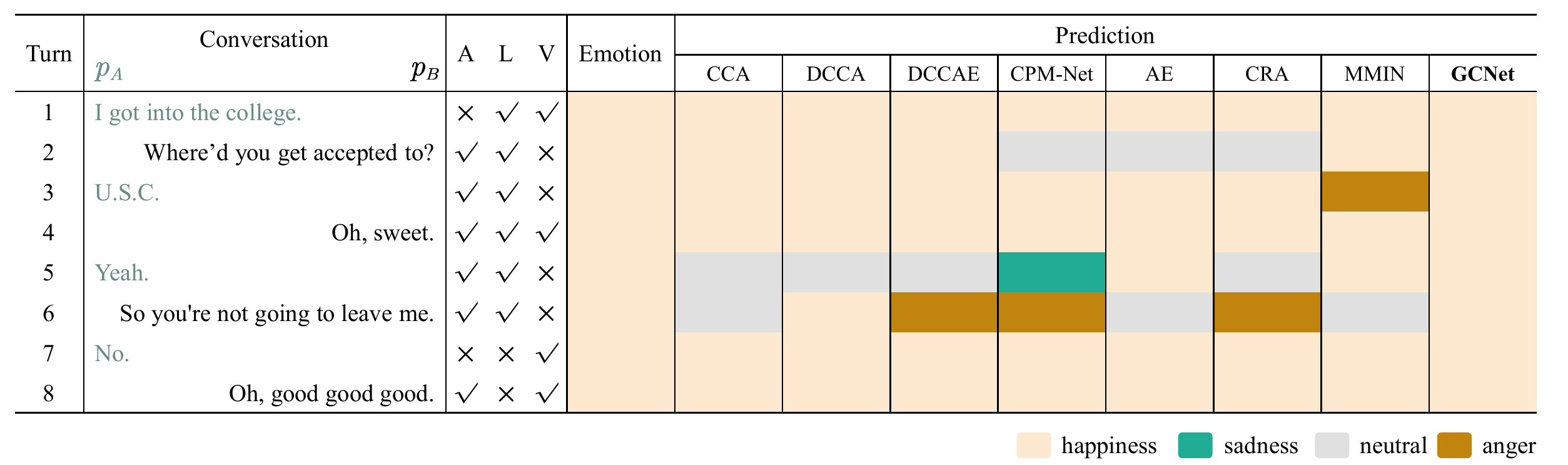}
		\caption{Prediction results on incomplete conversational data. ``$\surd$'' denotes the available modalities and ``$\times$'' denotes the missing modalities.}
		\label{Figure9}
	\end{figure*}

	\subsection{Convergence Analysis}
	In this section, we investigate the convergence property on the IEMOCAP(four-class) dataset with the missing rate $\eta=0.3$. Fig. \ref{Figure6-1}$\sim$\ref{Figure6-2} display the loss curves of classification and imputation, respectively. Despite the slight fluctuation, the loss curves maintain a descending trend and finally converge. For each loss function, the loss curves on training and validation sets converge at close epochs. These results prove the generalization ability of our GCNet on unseen data. For each data subset, the classification and imputation loss curves also converge at close epochs. These results reveal the correlation between classification and imputation. Our GCNet can achieve better classification performance when it can well reconstruct multimodal data.

	\subsection{Visualization of Embedding Space}
	To qualitatively analyze the improvements of GCNet, we visualize the latent representation of different methods on the IEMOCAP(four-class) test set. In the following experiments, we fix the missing rate $\eta$ to 0.3. 
	
	Fig. \ref{Figure7-1}$\sim$\ref{Figure7-7} present the t-SNE \cite{van2008visualizing} visualization results of baselines. Fig. \ref{Figure7-8} presents the visualization result of our GCNet. As can be observed, our proposed method can effectively disentangle the latent representation, making the margin between different classes clearer. Meanwhile, we visualize the latent representation of our method with increasing training iterations. Fig. \ref{Figure8-1}$\sim$\ref{Figure8-4} present the t-SNE results of the $10^{th}$, $20^{th}$, $50^{th}$ and $78^{th}$ epochs, respectively. As can be observed, with the increase of the epoch, the latent representation reveals the underlying class distribution much better.

	\subsection{Case Study}
	In this section, we compare the prediction results of different methods on incomplete conversational data. Fig. \ref{Figure9} provides a representative example from the IEMOCAP(four-class) test set. In this example, $p_A$ is accepted to a college and shares his happiness with $p_B$. We observe that GCNet generates more accurate results than other methods. Indeed, it is hard to make correct predictions on incomplete conversational data without incorporating temporal and speaker information. These qualitative results verify the effectiveness of our GCNet in incomplete multimodal learning.

	\section{Conclusions}
	\label{sec6}
	In this paper, we propose a novel framework, GCNet, for incomplete multimodal learning in conversations. It takes full advantage of speaker and temporal information in conversations, aiming to learn discriminative representations from modality-incomplete conversational data. Unlike existing works that mainly focus on medical images or individual utterances, we study incomplete multimodal learning on conversational data, filling the gap of current works. Experimental results on three benchmark datasets demonstrate the effectiveness of our method. Through quantitative and qualitative analysis, we first verify that GCNet consistently outperforms currently advanced approaches under varying missing rates, achieving the best classification and imputation performance. Then, we show the importance of incomplete data in feature learning and prove the necessity of each component in GCNet. After that, we visualize the parameter tuning process and reveal the impact of different hyper-parameters. Through convergence analysis, we also present the correlation between classification and imputation. Our GCNet can achieve better classification performance when it can well reconstruct modality-complete data.

	In the future, we will extend the applications of our proposed method. Besides conversational emotion recognition, we will leverage GCNet to deal with modality missing problems in other types of conversation understanding tasks. Furthermore, in addition to fixing the context window size to limit the number of edges, we will explore some dynamic edge construction strategies based on correlations between utterances in our future work.

	\section*{Acknowledgements}
	This work is founded by the National Natural Science Foundation of China (NSFC) under Grants 62201572, 61831022, 62276259 and U21B2010, and Open Research Projects of Zhejiang Lab under Grants 2021KH0AB06.
	
	\bibliographystyle{IEEEtran}
	\bibliography{mybib}
	
	\begin{IEEEbiography}[{\includegraphics[width=1.1in,height=1.25in,clip,keepaspectratio]{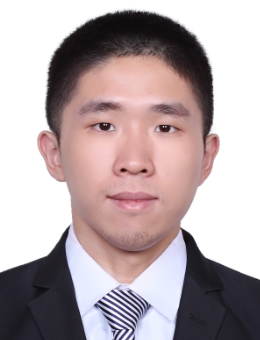}}]{Zheng Lian}
		received the B.S. degree from the Beijing University of Posts and Telecommunications, Beijing, China, in 2016. And he received the Ph.D degree from the Institute of Automation, Chinese Academy of Sciences, Beijing, China, in 2021. He is currently an Assistant Professor at National Laboratory of Pattern Recognition, Institute of Automation, Chinese Academy of Sciences, Beijing, China. His current research interests include affective computing, deep learning and multimodal emotion recognition.
	\end{IEEEbiography}
	\begin{IEEEbiography}[{\includegraphics[width=1.1in,height=1.25in,clip,keepaspectratio]{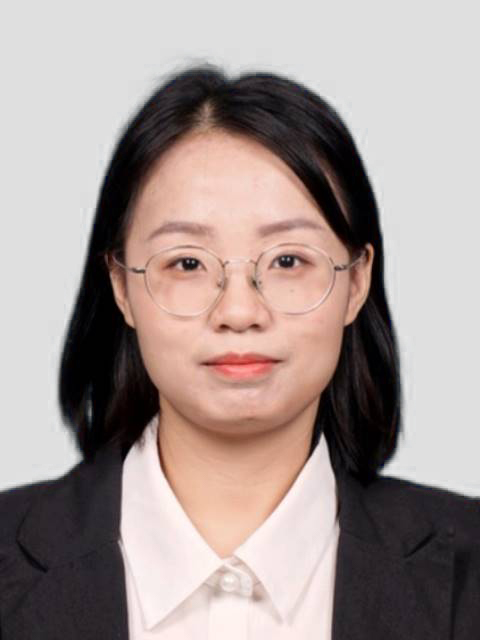}}]{Lan Chen}
		received the B.S. degree from the China University of Petroleum, Beijing, China, in 2016. And she received the Ph.D degree from the Institute of Automation, Chinese Academy of Sciences, Beijing, China, in 2022. Her current research interests include computer graphics and image processing.
	\end{IEEEbiography}
	\begin{IEEEbiography}[{\includegraphics[width=1.1in,height=1.25in,clip,keepaspectratio]{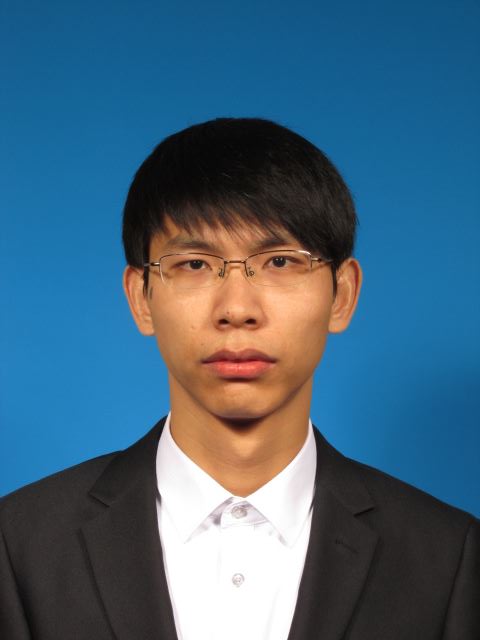}}]{Licai Sun}
		received the B.S. degree from Beijing Forestry University, Beijing, China, in 2016, and the M.S. degree from University of Chinese Academy of Sciences, Beijing, China, in 2019. He is currently working toward the Ph.D degree with the School of Artificial Intelligence, University of Chinese Academy of Sciences, Beijing, China. His current research interests include affective computing, deep learning and multimodal representation learning.
	\end{IEEEbiography}
	\begin{IEEEbiography}[{\includegraphics[width=1in,height=1.25in,clip]{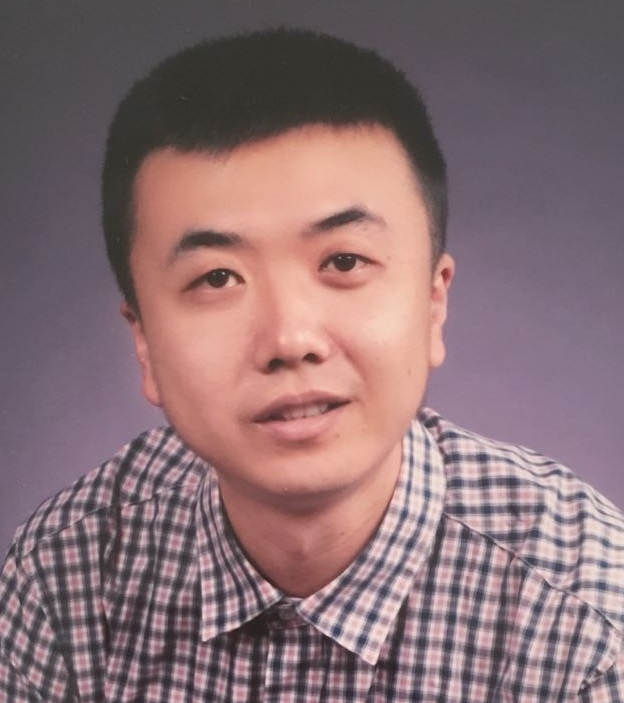}}]{Bin Liu}
		received his the B.S. degree and the M.S. degree from Beijing institute of technology, Beijing, China, in 2007 and 2009 respectively. He received Ph.D. degree from the National Laboratory of Pattern Recognition, Institute of Automation, Chinese Academy of Sciences, Beijing, China, in 2015. He is currently an Associate Professor in the National Laboratory of Pattern Recognition, Institute of Automation, Chinese Academy of Sciences, Beijing, China. His current research interests include affective computing and audio signal processing.
	\end{IEEEbiography}
	\begin{IEEEbiography}[{\includegraphics[width=1.1in,height=1.25in,clip,keepaspectratio]{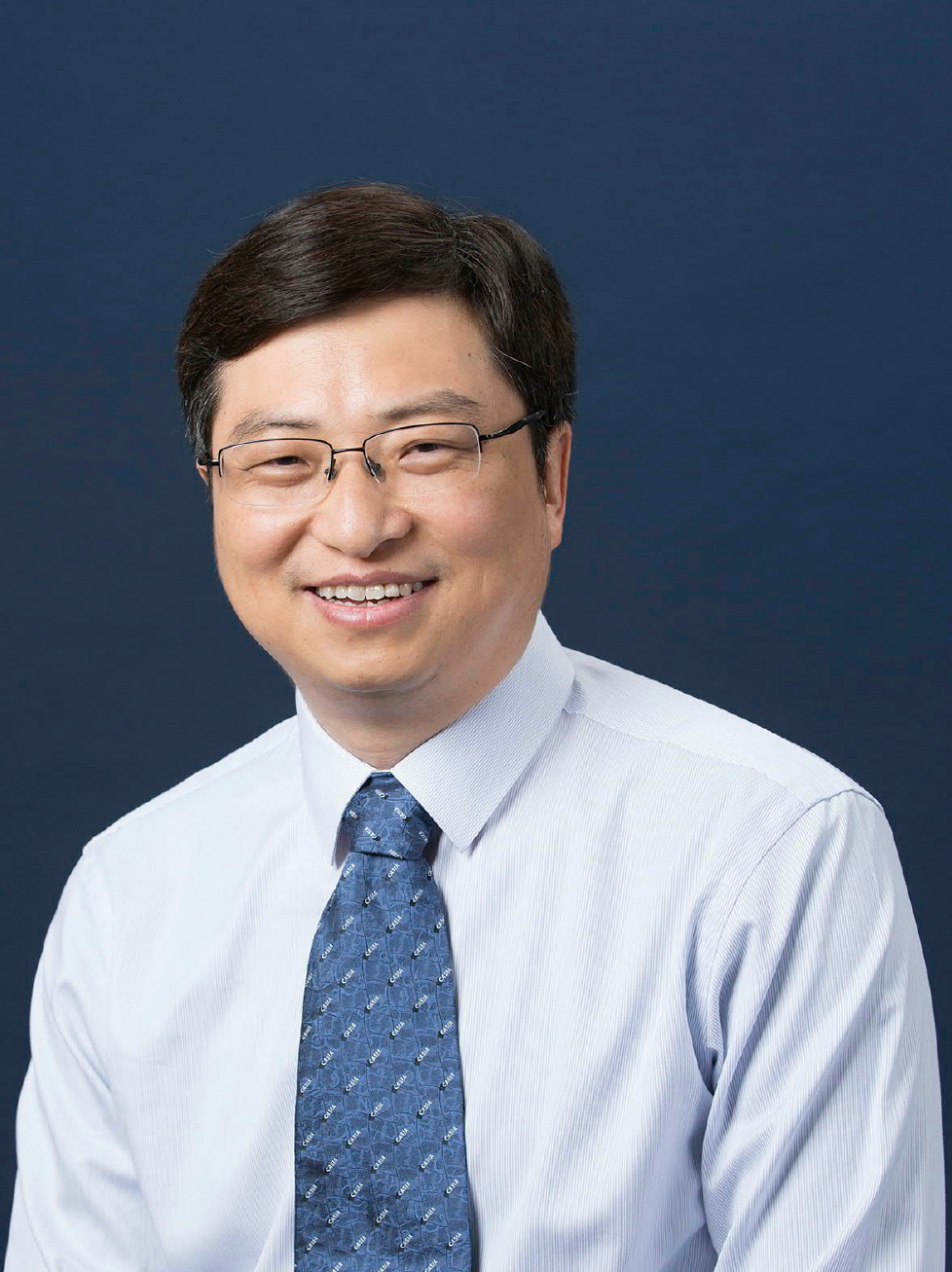}}]{Jianhua Tao}
		received the Ph.D. degree from Tsinghua University, Beijing, China, in 2001, and the M.S. degree from Nanjing University, Nanjing, China, in 1996. He is currently a Professor with Department of Automation, Tsinghua University, Beijing, China. He has authored or coauthored more than eighty papers on major journals and proceedings. His current research interests include speech recognition, speech synthesis and coding methods, human–computer interaction, multimedia information processing, and pattern recognition. He is the Chair or Program Committee Member for several major conferences, including ICPR, ACII, ICMI, ISCSLP, NCMMSC, etc. He is also the Steering Committee Member for the IEEE Transactions on Affective Computing, an Associate Editor for Journal on Multimodal User Interface and International Journal on Synthetic Emotions, and the Deputy Editor-in-Chief for Chinese Journal of Phonetics. He was the recipient of several awards from the important conferences, such as Eurospeech, NCMMSC, etc.
	\end{IEEEbiography}
	
\end{document}